%% file: PRN.tex
\input{_constants}
\arxiv 

\pdfoutput=1
\documentclass[10pt,twocolumn,letterpaper]{article}
\input{cvpr_header}
\usepackage{graphicx}
\usepackage{booktabs}
\usepackage{amssymb}
\usepackage[accsupp]{axessibility}

\newcommand\blfootnote[1]{%
  \begingroup
  \renewcommand\thefootnote{}\footnote{#1}%
  \addtocounter{footnote}{-1}%
  \endgroup
}

\makeatletter

\newcommand{\Rmnum}[1]{\expandafter\@slowromancap\romannumeral #1@}
\makeatother
\begin{document}
\title{\paperTitle}
\author{\authorBlock}
\maketitle

\begin{abstract}
\blfootnote{$^{*}$ Corresponding author.}
Anomaly detection and localization are widely used in industrial manufacturing for its efficiency and effectiveness.
Anomalies are rare and hard to collect and supervised models easily over-fit to these seen anomalies with a handful of abnormal samples, producing unsatisfactory performance. 
On the other hand, anomalies are typically subtle, hard to discern, and of various appearance, making it difficult to detect anomalies and let alone locate anomalous regions.
To address these issues, we propose a framework called Prototypical Residual Network (PRN), which learns feature residuals of varying scales and sizes between anomalous and normal patterns to accurately reconstruct the segmentation maps of anomalous regions.
PRN mainly consists of two parts: multi-scale prototypes that explicitly represent the residual features of anomalies to normal patterns; a multi-size self-attention mechanism that enables variable-sized anomalous feature learning.
Besides, we present a variety of anomaly generation strategies that consider both seen and unseen appearance variance to enlarge and diversify anomalies.
Extensive experiments on the challenging and widely used MVTec AD benchmark show that PRN outperforms current state-of-the-art unsupervised and supervised methods. We further report SOTA results on three additional datasets to demonstrate the effectiveness and generalizability of PRN.
\end{abstract}

\section{Introduction}
\label{sec:intro}

\begin{figure}[t]
  \centering
   \includegraphics[width=1.0\linewidth]{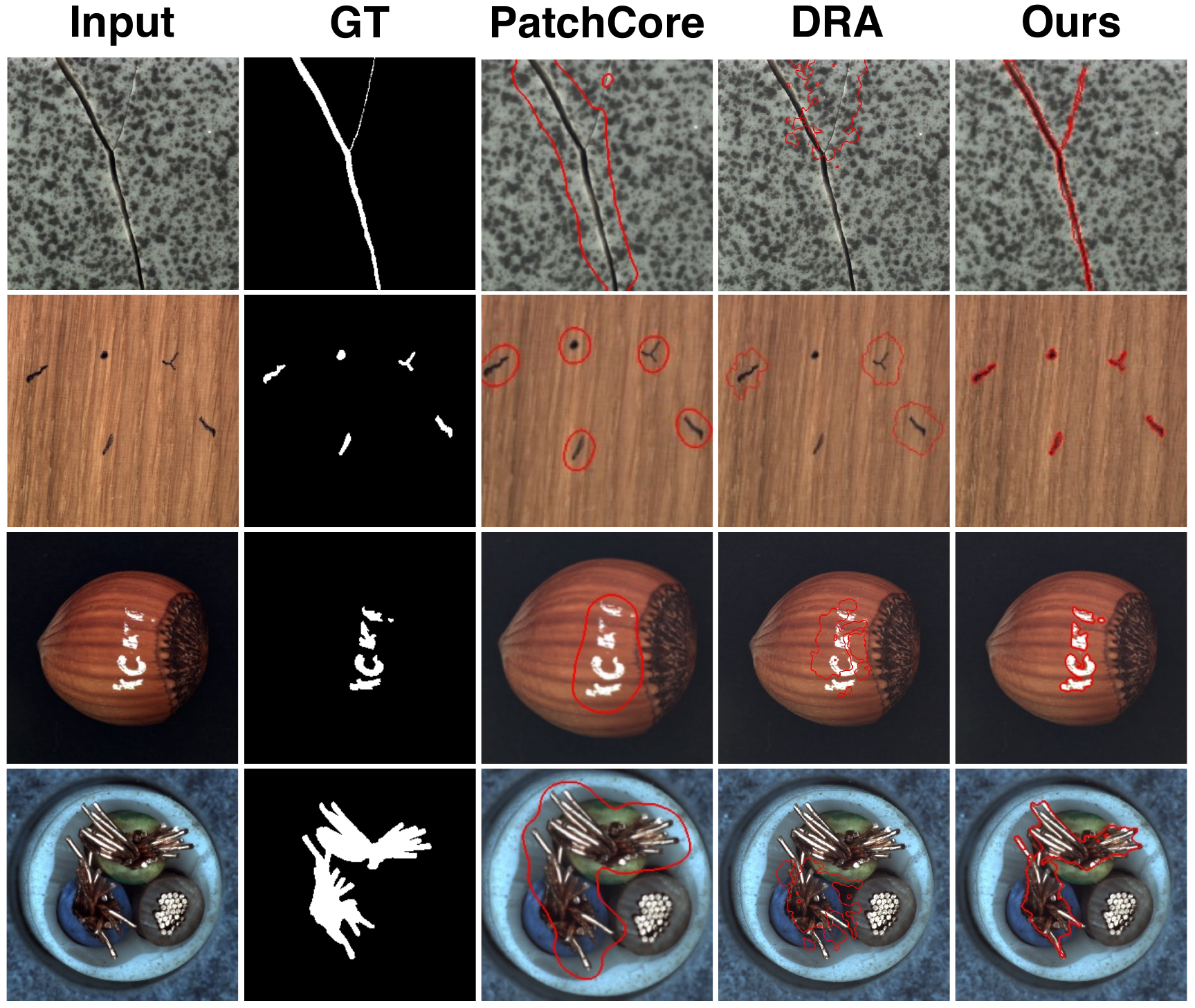}

   \caption{Anomaly detection and localization examples on MVTec\cite{bergmann2019mvtec}. Compared with the unsupervised method PatchCore\cite{roth2022patchcore} and the supervised method DRA\cite{ding2022dra}, the proposed PRN is able to locate the anomalous regions more accurately.}
   \label{fig:visualization results}
\end{figure} 

The human cognition and visual system has an inherent ability to perceive anomalies \cite{tao2022al_survey}. Not only can humans distinguish between defective and non-defective images, but they can also point to the location of anomalies even if they have seen none or only a limited number of anomalies. 
Anomaly detection (image-level binary classification) and anomaly localization (pixel-level binary classification) are introduced for the same purpose, and have been widely used in various scenarios due to their efficiency and remarkable accuracy, including industrial defect detection\cite{bergmann2019mvtec,wieler2007dagm,mishra2021btad,bozic2021SDD2}, medical image analysis\cite{seebock2016medical} and video surveillance\cite{liu2018video}.

Given its importance, a significant amount of work has been devoted to anomaly detection and anomaly localization, but few have addressed both detection and localization problems well at the same time.
We argue that real-world anomalous data weaken these models mainly in three aspects:
\Rmnum{1}) the amount of abnormal samples is limited and significant fewer than normal samples, producing data distributions that lead to a naturally \textbf{imbalanced learning} problem;
\Rmnum{2}) anomalies are typically subtle and hard to discern, since normal patterns still dominate the anomalous image; \textbf{identifying abnormal regions} out of the whole image is the key to anomaly detection and localization;
\Rmnum{3}) the appearance of anomalies varies significantly, \ie, abnormal regions can take on a variety of sizes, shapes and numbers, and such \textbf{appearance variations} make it challenging to well-localizing all the anomalies.

Without adequate anomalies for training, unsupervised models become the de facto dominant approaches, which get rid of the imbalance problem by learning the distribution of normal samples\cite{rudolph2021differnet,gudovskiy2022cflow,rudolph2022cs-flow,yu2021fastflow,lee2022cfa,bergmann2020us,defard2021padim,cohen2020spade,salehi2021kdad,deng2022rd,roth2022patchcore} or generating sufficient synthetic anomalies\cite{li2021cutpaste,zavrtanik2021draem,yang2022memseg,schluter2022nsa,liang2022ocr-gan}. 
However, these methods are opaque to genuine anomalies, resulting in implicit decisions that may induce many false negatives and false positives. 
Besides, unsupervised methods rely heavily on the quality of normal samples, and thus are not robust enough and perform poorly on uncalibrated or noisy datasets~\cite{han2022adbench}.
As shown in \cref{fig:visualization results}, unsupervised models predict broad regions around the anomaly. We attribute this problem to less discriminative abilities of these methods.

\begin{figure}[t]
  \centering
   \includegraphics[width=1.0\linewidth]{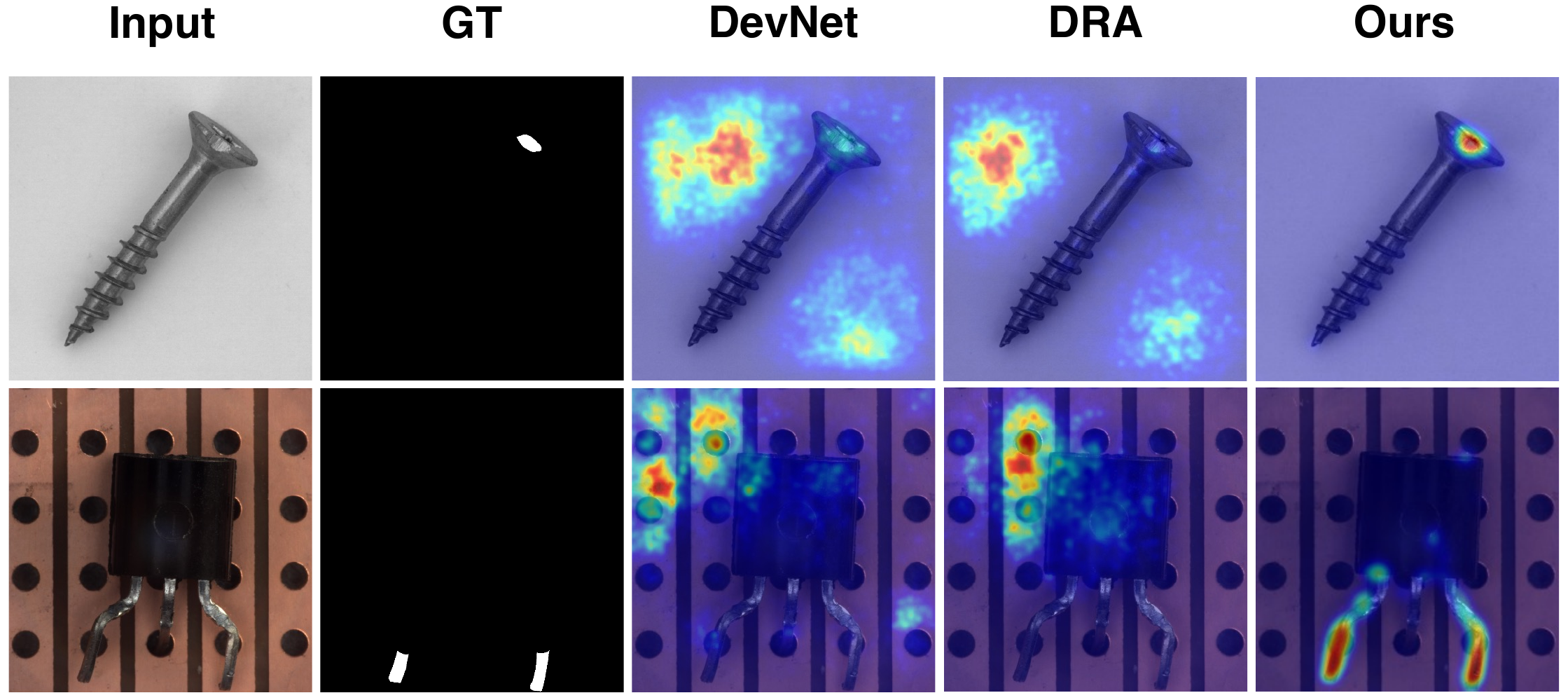}
   \caption{Indecipherable problem of supervised methods DevNet\cite{pang2021devnet} and DRA\cite{ding2022dra}. Both images are detected as anomalous. Other methods mistakenly highlight normal regions rather than defect regions, whereas PRN correctly pinpoints the defect regions.}
   \label{fig:uninterpretable}
\end{figure}

Recently, several supervised methods~\cite{pang2021devnet,ding2022dra,ruff2019deep-sad} are introduced. 
DeepSAD~\cite{ruff2019deep-sad} enlarges the margin between the anomaly and the one-class center in the latent space to obtain more compact one-class descriptors by limit seen anomalies.
DRA~\cite{ding2022dra} and DevNet~\cite{pang2021devnet} formulate anomaly detection as a multi-instance learning (MIL) problem, scoring an image as anomaly if any image patch is a defect region. MIL-based methods enforce the learning at fine-grained image patch
level, which effectively reduces the interference of normal patches in the anomalous images. Yet, these approaches typically struggle to accurately locate all anomalous regions with image-level supervision, as shown in \cref{fig:visualization results}. In particular, when the anomalous regions only occupy a tiny part of image patches, image-level representation may be dominated by the normal regions and disregards tiny anomalous, which may cause inconsistent image-level and pixel level performance as shown in \cref{tab:image and pixel-auroc}. Furthermore, as shown in \cref{fig:uninterpretable}, these methods also encounter uninterpretable problems when making decisions.

In this paper, we propose a framework called Prototypical Residual Network (PRN) as an effective remedy for aforesaid issues on anomaly detection and localization. First, we propose multi-scale prototypes to represent normal patterns.
In contrast to previous methods for constructing normal patterns from concatenated feature memory~\cite{roth2022patchcore} or random sampled feature maps~\cite{yang2022memseg}, we construct normal patterns with prototypes of intermediate feature maps of different scales, thereby preserving the spatial information and providing precise and representative normal patterns. 
Further, we obtain the feature map residuals via the deviation between the anomalous image and the closest prototype at each scale, and we add multi-scale fusion blocks to exchange information across different scales.
Second, since the appearance of anomaly regions varies a lot, it is necessary to learn relationships among patches from multiple receptive fields. Thus, we introduce a multi-size self-attention\cite{wang2022m2tr,vaswani2017attention,meng2022adavit,wang2022bevt} mechanism, which operates on patches of different receptive fields to detect patch-level inconsistencies at different sizes. 
Finally, unlike previous methods~\cite{ding2022dra,pang2021devnet} that use image-level supervision for training, our model learns to reconstruct the anomaly segmentation map with pixel-level supervision, which focuses more on the anomalous regions and preserves better generalization. 
Besides, we put forward a variety of anomaly generation strategies that efficiently mitigate the impact of data imbalance and enrich the anomaly appearance.
With the proposed modules, our method achieves more accurate localization than previous unsupervised and supervised methods, as shown in \cref{fig:visualization results} and \cref{fig:uninterpretable}.

The main contributions of this paper are summarized as follows:
\begin{itemize}
\item We propose a novel Prototypical Residual Networks for anomaly detection and localization. Equipped with multi-scale prototypes and the multi-size self-attention mechanism, PRN learns residual representations among multi-scale feature maps and within the multi-size receptive fields at each scale.
\item We present a variety of anomaly generation strategies that considering both seen and unseen appearance variance to enlarge and diversify anomalies.
\item We perform extensive experiments on four datasets to show that our approach achieves new SOTA anomaly detection performance and outperforms current SOTA in anomaly localization performance by a large margin.
\end{itemize}

\section{Related Work}
\label{sec:related}

\textbf{Unsupervised Approaches.\quad}Unsupervised paradigm assumes that only normal data is available during training\cite{pang2021survey,ruff2021survey,tao2022al_survey}. Auto-Encoder based methods\cite{youkachen2019cae,akcay2019skip,gong2019mem-ae,bergmann2018ssim} rely on the hypothesis that the model is trained to reconstruct normal regions well but fails for abnormal regions. Although localization results based on the difference between the input and the reconstructed image are often intuitive and interpretable, their performance is limited.
Generative models are introduced to obtain better reconstruction performance. However, the generation effect of VAE\cite{dehaene2020gvae,liu2020avae,dehaene2020gvae} or GAN\cite{schlegl2017gan,hou2021divide-gan,akcay2018ganomaly,li2020cgan,schlegl2019fgan,yu2020cyclegan} over normal areas in the image is poor, leading to coarse reconstruction and false detection. Normalizing flows based methods\cite{rudolph2021differnet,gudovskiy2022cflow,rudolph2022cs-flow,yu2021fastflow} learn bijective transformations between data distributions and well-defined densities, however the computational cost of these approaches is significant.  Knowledge distillation-based methods\cite{bergmann2020us,salehi2021kdad,wang2021stpm,deng2022rd} transform the anomaly detection task into a feature comparison between teacher and student networks. Deep feature modeling-based methods\cite{cohen2020spade,defard2021padim,kim2021semi-orthogonal,mishra2021btad,roth2022patchcore,zheng2022fyd,lee2022cfa} bulid a feature space for input images and then detect and localize anomalies by comparing the features. Self-supervised learning-based methods\cite{yi2020patchsvdd,pirnay2022intra,ristea2022sspcab} designed proxy tasks such as predicting or recovering hidden regions or properties in input images\cite{tao2022al_survey}. One-class classification based methods\cite{scholkopf2001ocsvm,ruff2018deep-svdd,yi2020patchsvdd} aim to map training images or patches to a small hypersphere in the feature space. These approaches address the imbalance problem by being opaque to anomalous samples, but suffer from implicit decisions that result in subpar performance on subtle and challenging anomalies.

\textbf{Supervised Approaches.\quad}A recent emerging trend focuses on supervised anomaly detection by leverging seen anomalies to increase the differentiation between anomalous and normal samples.
Some existing works\cite{gornitz2013sad,liu2019mlep,ruff2019deep-sad} are learned with a minority of anomaly based on one-class classification metric. Some anomaly-focused deviation losses proposed in \cite{zhang2020viral,pang2021devnet} mitigate the bias derived from the seen anomalies. A multi-head model is introduced in \cite{ding2022dra} to learn disentangled anomaly representations, where each head is dedicated to capturing a specific type of anomaly. Due to the imbalanced learning problem, these methods are prone to over-fitting to seen anomalies and fail to generalize to unseen anomalies, resulting in poor anomaly detection performance. In addition, image-level representations may be dominated by normal regions while disregarding the representations of subtle anomaly regions, resulting in an inability to accurately localize anomalies that come in a variety of sizes, shapes and numbers.

\section{Method}
\label{sec:method}

\begin{figure*}
  \centering
  \includegraphics[width=0.8\linewidth]{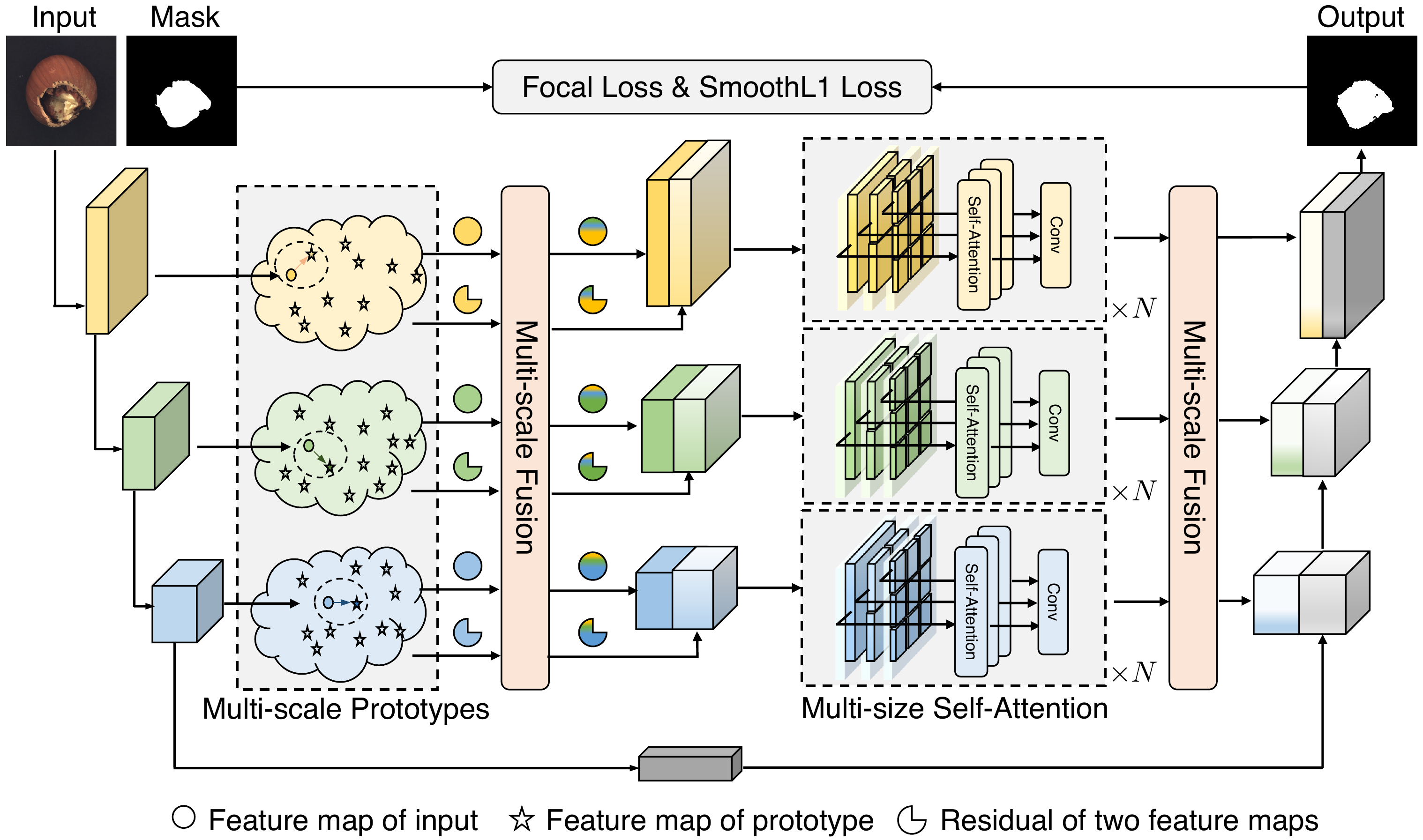}
  \caption{An overview of the proposed Prototypical Residual Network. Anomalous feature residuals of inputs are obtained via Multi-scale Prototypes for each scale feature map from the nearest cluster prototype. Feature maps and residuals at different scales are separately fused by Multi-scale Fusion blocks. Multi-size Self-Attention learns feature residuals on patches of different sizes at each scale, which are further enhanced by another Multi-scale Fusion block. Please see text for details.}
  \label{fig:architecture}
\end{figure*}

Together with the proposed anomaly generation strategies (\cref{subsec:anomaly generation strategies}) that retain a balanced data distribution, we propose the Prototypical Residual Network (PRN) to reconstruct the segmentation map for anomaly detection and localization. Overall, we adopt a U-Net~\cite{ronneberger2015U-Net}-like architecture as shown in \cref{fig:architecture}. The encoder is a pre-trained ResNet-18\cite{he2016resnet-18}, and the decoder consists of upsampling and convolution blocks. The skip-connection branches of PRN are equipped with the proposed Multi-scale Prototypes (MP, \cref{subsec:multi-scale prototypes}), Multi-scale Fusion blocks (MF, \cref{subsec:multi-scale fusion}) and a Multi-size Self-Attention mechanism (MSA, \cref{subsec:multi-size self-attention}).
In the following, we will concretely describe each part.

\subsection{Multi-scale Prototypes}
\label{subsec:multi-scale prototypes}
\textbf{Prototype Initialization.\quad}We define $\mathcal{X}_{N}$ to be the set of all normal samples during training ($\forall x \in \mathcal{X}_{N}: y_{x} = 0$).
$y_{x}$ denotes that if an image $x$ is normal (0) or abnormal (1). Following \cite{roth2022patchcore,defard2021padim}, we use a network pre-trained on ImageNet to obtain feature maps of the input image at different scales. We use $\mathcal{F}_{i, j}=\mathcal{F}_{j}\left(x_{i}\right)$ ($j \in\{1,2,3,4\}$) to denote the $j$-th block output of input $x_{i}$ from a ResNet-like architecture such as  ResNet-18~\cite{he2016resnet-18}. Assume the feature map $\mathcal{F}_{i, j}\in \mathbb{R}^{c^{j} \times h^{j} \times w^{j}}$ to be a tensor of depth $c^{j}$, height $h^{j}$ and width $w^{j}$. Firstly, the $j$-th scale prototypes $\mathcal{P}_{j}\in \mathbb{R}^{K\times c^{j} \times h^{j} \times w^{j}}$ are $K$ feature maps randomly sampled from $\mathcal{F}_{j}\left(\mathcal{X}_{N}\right)$, and are updated by k-means clustering\cite{hartigan1979k-means}. L2 distance is used to calculate the distance between two feature maps. As the number of normal samples in different datasets varies considerably, to have a suitable amount of prototypes, we set the number of prototypes to a certain ratio of the number of normal samples.
As a result, the value of $K$ varies by datasets. The ablation on the proportion number is detailed in \cref{subsec:Ablation Study}, and is typically $10\%$. Three scales of prototypes are employed ($j \in\{1,2,3\}$). Model parameters are frozen during clustering. After clustering, the prototypes $\mathcal{P}_{j}\in \mathbb{R}^{K\times c^{j} \times h^{j} \times w^{j}}$ at each scale remain unchanged during subsequent model training.

\textbf{Residual Representation.\quad} Given the $i$-th input image and its corresponding feature map $\mathcal{F}_{i, j}$ at $j$-th block, we can find the closest prototype $\mathcal{P}_j^*$ at $j$-th scale by calculating the L2 distance between $\mathcal{F}_{i, j}$ and each of the prototypes $\mathcal{P}_j$. 
We define the anomalous residual representation of $\mathcal{F}_{i,j}$ to its closest prototype as
\begin{equation}
\begin{aligned}
    \mathcal{D}_{i,j}=&D(\mathcal{F}_{i,j} -\mathcal{P}_j^*), \\
    \text{s.t.}\  \mathcal{P}_j^*=&\underset{\mathcal{P}_j^k \subset \mathcal{P}_j}{\arg \min } \|\mathcal{F}_{i,j} -\mathcal{P}_j^{k}\|_{2}
\end{aligned}
\end{equation}
where $D(\cdot, \cdot)$ implements the element-wise Euclidean distance between two tensors, 
$\mathcal{D}_{i,j}\in \mathbb{R}^{c^{j} \times h^{j} \times w^{j}}$ is the residual from the nearest cluster prototype $\mathcal{P}_j^*$. Note that the input sample can match distinct prototypes at different scales, as the prototypes are learned independently at each scale.

\begin{figure}
  \centering
   \includegraphics[width=1.0\linewidth]{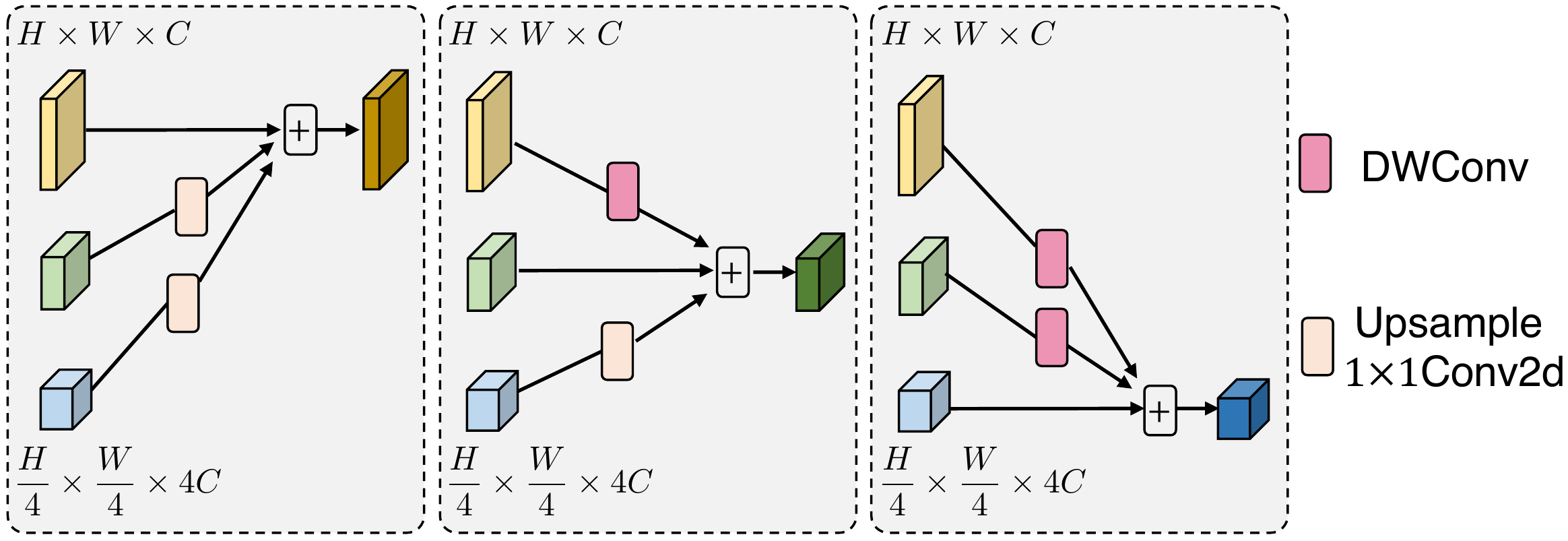}
   \caption{Three feature maps of different scales are fused by a multi-scale fusion block.}
   \label{fig:multi-scale fusion}
\end{figure}

\subsection{Multi-scale Fusion}
\label{subsec:multi-scale fusion}

To enable information exchanging across multi-scale representations, we propose to use Multi-scale Fusion blocks (MF) inspired by \cite{gu2022hrvit,wang2020hrnet}. As shown in \cref{fig:multi-scale fusion}, the fused output feature map is the sum of the transformed representations of three input feature maps. The feature map $\mathcal{F}_{i, j}^*$ is fused with others as follows:  
\begin{equation}
    \mathcal{F}_{i,j}^{*}=f_{1 j}\left(\mathcal{F}_{i,1}\right)+f_{2 j}\left(\mathcal{F}_{i,2}\right)+f_{3 j}\left(\mathcal{F}_{i,3}\right)
    \label{eq:multi-scale fusion}
\end{equation}
The choice of the transform function $f_{r j}(\cdot)$ depends on the input feature map index r and the output feature map index j ($r, j \in\{1,2,3\}$). If $r=j$, $f_{r j}(\mathcal{F}_{i,r})=\mathcal{F}_{i,r}$.  If $r<j$, $f_{r j}(\mathcal{F}_{i,r})$ down-samples the input feature map $\mathcal{F}_{i,r}$ through depth-wise separable convolutions with a stride of $2^{j-r}$, a kernel size of $2^{j-r}+1$ and a padding of $2^{j-r-1}$. If $r>j$, $f_{r j}(\mathcal{F}_{i,r})$ up-samples the input feature map $\mathcal{F}_{i,r}$ through a bilinear up-sampling followed by a $1 \times 1$ convolution. The anomalous residual representation $\mathcal{D}_{i,j}$ also follows the fusion paradigm in \cref{fig:multi-scale fusion} and is concated with $\mathcal{F}_{i,j}^{*}$ along the depth dimension to obtain $\mathcal{C}_{i,j}^{*}\in \mathbb{R}^{2c^{j} \times h^{j} \times w^{j}}$ .

\subsection{Multi-size Self-Attention}

\label{subsec:multi-size self-attention}

As the anomalous regions vary in magnitude, to further detect local inconsistencies in the concatenated feature maps $\mathcal{C}_{i,j}^{*}$, we introduce a Multi-size Self-attention (MSA) mechanism. MSA splits $\mathcal{C}_{i,j}^{*}$ into patches of different sizes $p_{s}\in\{h^j,h^j/2,h^j/4,h^j/8\}$ and computes patch-wise self-attention\cite{wang2022m2tr,vaswani2017attention,weng2022transformer,tian2022resformer,xing2022svformer} in different heads. Different heads correspond to different patch sizes, as shown in \cref{fig:architecture}. To be specific, we first extract patches of shape $2c^j \times p_{s} \times p_{s}$  from $\mathcal{C}_{i,j}^{*}$, and flatten them into 1-dimension vectors for the $s$-th head. And then we use fully-connected layers to embed the flattened vectors into query embeddings $\mathcal{Q}_{i,j}^{s} \in \mathbb{R}^{\mathcal{N} \times c^s}$, where $\mathcal{N}=\left(h^j /  p_{s}\right) \times\left(w^j /  p_{s}\right)$ and $c^{s}=2c^j \times p_s \times p_s$. We obtain key embeddings $\mathcal{K}_{i,j}^{s}$ and value embeddings $\mathcal{V}_{i,j}^{s}$ with the similar operations. The attention matrix is calculated by the following process:
\begin{equation}
    \mathcal{A}_{i,j}^{s}=\operatorname{softmax}\left(\frac{\mathcal{Q}_{i,j}^{s}\left(\mathcal{K}_{i,j}^{s}\right)^{T}}{c^s}\right) \mathcal{V}_{i,j}^{s} 
    \label{eq:attention}
\end{equation}
After that, $\mathcal{A}_{i,j}^{s}$ is reshaped to the original spatial resolution. Similar operations are implemented to obtain features from heads of different patch sizes. Finally, these features are concatenated and passed through a 2D residual block to obtain the output $\mathcal{T}_{i,j}\in \mathbb{R}^{2c^{j} \times h^{j} \times w^{j}}$. We stack the MSA for N times (N = 3 in this paper). To further fuse multiple scales of $\mathcal{T}_{i,j}$, we use 
another MF block to obtain $\mathcal{T}_{i,j}^*$, which is the output of the skip-connection as shown in \cref{fig:architecture}.

\begin{figure}
  \centering
  \includegraphics[width=1.0\linewidth]{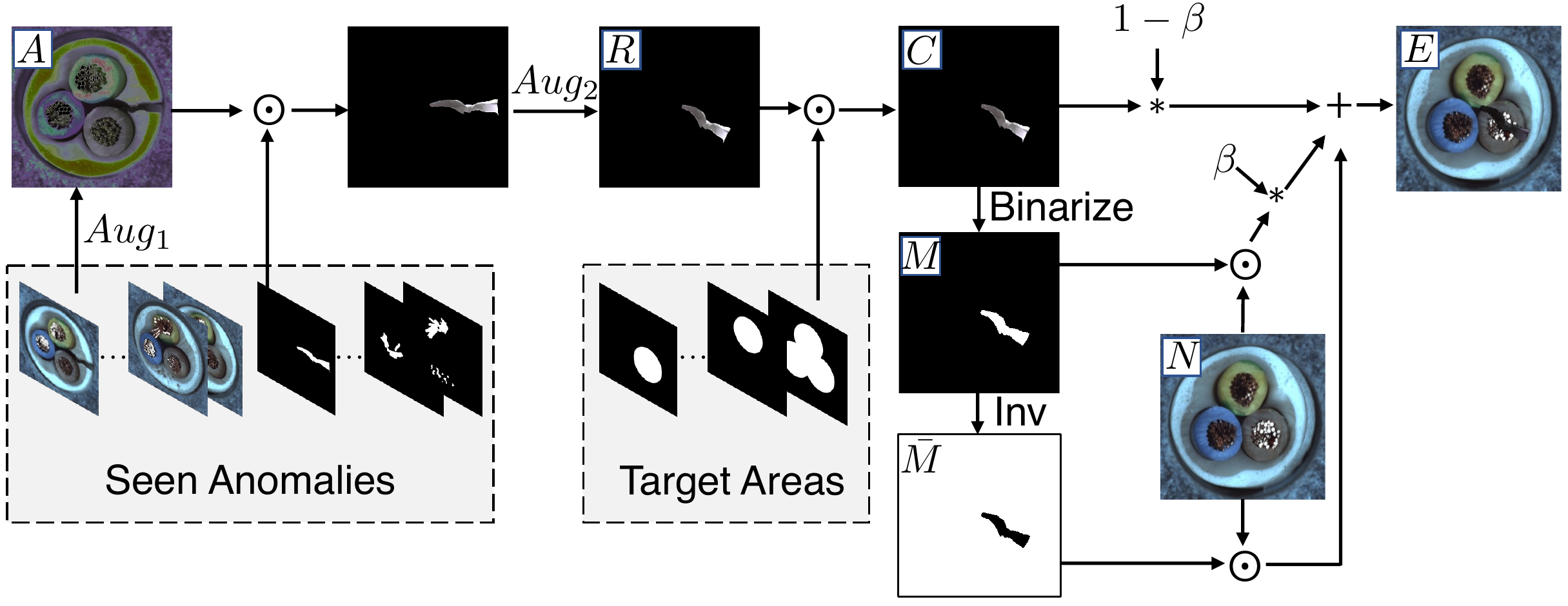}
  \caption{Generating extended anomalies. The anomalous region is augmented by random augmentation and placed on a target area of the normal sample to generate various anomalies online.}
  \label{fig:extended anomalies} 
\end{figure}

\subsection{Anomaly Generation Strategies}
\label{subsec:anomaly generation strategies}

To alleviate the data imbalance problem, we propose two kinds of online anomaly generation strategies that can generate various types of anomalies. One strategy is to create in-distribution anomalies by placing augmented anomalous regions from seen anomalies on normal samples, and these generated anomalies are named extended anomalies (EA). EA enlarge the 
amount of anomalies and mitigate the data imbalance problem. Another strategy is to create out-of-distribution anomalies\cite{zavrtanik2021draem} using normal samples without knowledge of the seen anomalies. These generated anomalies are named simulated anomalies (SA) , which supplement potential unseen anomalies.

\textbf{Extended Anomalies.\quad}Instead of simply augmenting the entire image from the seen anomalies, we augment the specific anomalous regions of the seen anomalies and place them at any possible position within the normal sample. First, augmentations (\cref{fig:extended anomalies}, $Aug_1$) are applied to a randomly selected anomaly from the seen anomalies in order to generate color varieties (\cref{fig:extended anomalies}, $A$). $Aug_1$ takes two random operations from \{ equalize, solarize, posterize, sharpness, auto-contrast, invert, gamma-contrast \}. After that, we argument the selected anomaly with random spatial transformations as \{ rotate, shear, shift \} to obtain position and shape varieties (\cref{fig:extended anomalies}, $R$). Since the extended anomalies should be as realistic and reasonable as possible, we propose a soft position constrain to place $R$ in the foreground. More specifically, Target Areas (TA) is used to refer to areas where anomalies can be placed. We crop $R$, using a randomly sampled target area, to obtain clipped anomaly region (\cref{fig:extended anomalies}, $C$). If $R$ has no overlap with the target region, we perform $Aug_2$ again until $R$ has overlap with the target region. We binarized $C$ to obtain the ground truth mask (\cref{fig:extended anomalies}, $M$). The proposed extended anomalies (\cref{fig:extended anomalies}, $E$) is therefore defined as:
\begin{equation}  
	E=\bar{M} \odot N+(1-\beta)C+\beta\left(M \odot N\right) 
\end{equation}
where $\bar{M}$ is the inverse of $M$, $\odot$ is the element-wise multiplication operation, $\beta$ is the opacity parameter\cite{zavrtanik2021draem} for better combination of abnormal and normal parts. For object datasets and texture datasets, the target areas are part of the foreground of the object and part of the whole image, respectively. The shapes of the target are the set of geometries:  \{circle, rectangular, polygonal\}.

\textbf{Simulated Anomalies.\quad} Similar to DRAEM\cite{zavrtanik2021draem}, we multiply Perlin\cite{perlin1985perlin} noise with random textures from the DTD\cite{cimpoi2014DTD} dataset and apply these augmented textures to normal images. As these anomalies significantly differs from the seen anomalies, we refer to these out-of-distribution anomalies as heterologous anomalies (HEA). To further expand the diversity of simulated anomalies, we introduce homologous anomalies (HOA), in which anomalies multiplied by the Perlin noise are augmented normal images. Note that the TA mentioned above is also applied to the generation of simulated anomalies. More details about HEA and HOA are presented in the supplementary materials. 

\subsection{Training and Inference}
\label{subsec:training and inference }

The decoder of PRN outputs an anomaly score map $\mathcal{M}_o$, which is of the same shape as the ground truth mask $\mathcal{M}$. Inspired by \cite{zavrtanik2021draem} and \cite{yang2022memseg} , a focal loss\cite{lin2017focal} and a smooth L1 loss\cite{girshick2015fast-rcnn} are applied to increase the robustness toward accurate segmentation of hard examples and reduce the over-sensitivity to outliers, respectively. Thus, the total loss $\mathcal{L}_{total} $ used for training PRN is defined as
\begin{equation}
    \mathcal{L}_{total} = \text{Smooth}_{\mathcal{L}1}\left(\mathcal{M}_o, \mathcal{M}\right)+\lambda\mathcal{L}_{focal}\left(\mathcal{M}_o, \mathcal{M}\right)
    \label{eq:loss}
\end{equation}
When the predicted $\mathcal{M}_o$ is accurate and sufficiently close to $\mathcal{M}$, $\mathcal{M}_o$ can be interpreted not only as the pixel-level anomaly localization result, but also as an image-level anomaly estimation for anomaly detection. Specifically, we take the average of the top-K anomalous pixels as the image-level anomaly score for anomaly detection. In a preliminary study, we trained a classification network based on $\mathcal{M}_o$ for image-level anomaly detection, but did not observe an improvement over top-K estimation.

\section{Experiments}
\label{sec:experiments}

\subsection{Experimental Details}
\label{subsec:Experimental Details}

\textbf{Datasets.\quad}To validate the effectiveness and generalizability of our approach, we perform experiments on various datasets, \ie, MVTec Anomaly Detection (MVTec AD\cite{bergmann2019mvtec}), DAGM\cite{wieler2007dagm}, BeanTech anomaly detection dataset (BTAD\cite{mishra2021btad}), and KolektorSDD2\cite{bozic2021SDD2}. There are 10 object sub-datasets and 5 texture sub-datasets in MVTec AD. Each sub-dataset presents a diverse set of anomalies, which enables a general evaluation of surface anomaly detection methods. DAGM contains 10 textured objects with small abnormal regions that are visually very similar to the background. BTAD includes three categories of real-world industrial products showcasing different body and surface defects. KolektorSDD2 is a dataset of surface defects that vary in shape, size, and color, from small scratches and spots to large surface defects. We adopt the general supervised setting\cite{ding2022dra,pang2021devnet}, where the training set of each sub-dataset contains only 10 abnormal samples. More details will be provided in the supplementary materials.

\textbf{Evaluation Metrics.\quad} Following previous work, we evaluate the results via the area under the receiver operating characteristic curve at the image level (Image-AUROC) and pixel level (Pixel-AUROC). However, anomalous regions typically only occupy a tiny fraction of the entire image. Thus, Pixel-AUROC can not accurately reflect the localization accuracy due to the fact that the false positive rate is dominated by the extremely high number of non-anomalous pixels and remains low despite false positive detection\cite{tao2022al_survey}. To comprehensively measure localization performance, we introduce Per Region Overlap (PRO)\cite{bergmann2020us} score and pixel-level Average Precision (AP)\cite{zavrtanik2021draem}. The PRO score treats anomaly regions of varying sizes equally\cite{roth2022patchcore,deng2022rd}, while AP is more appropriate for highly imbalanced classes, especially for industrial anomaly localization, where accuracy is critical\cite{zavrtanik2021draem}.

\begin{table*}[]
\centering
\scalebox{0.7}{
\begin{tabular}{@{}lcccccccccc@{}}
\toprule
\multirow{2}{*}{Category}     & \multicolumn{7}{c}{Unsupervised}                                               & \multicolumn{3}{c}{Supervised} \\
            \cmidrule(lr){2-8}\cmidrule(lr){9-11} 
           & KDAD\cite{salehi2021kdad}& CFLOW\cite{gudovskiy2022cflow} & DRAEM\cite{zavrtanik2021draem} & SSPCAB\cite{ristea2022sspcab} & CFA\cite{lee2022cfa} & RD4AD\cite{deng2022rd} & PatchCore\cite{roth2022patchcore}  & DevNet\cite{pang2021devnet} & DRA\cite{ding2022dra}  & Ours   \\ \midrule
Carpet     & 80.3/95.5 & 97.6/\textbf{99.2}  & 96.9/97.5  & 93.1/92.6  & 99.3/98.6  & 98.7/98.9  & 99.1/99.0  & 82.5/97.2   & 92.5/98.2   & \textbf{99.7}/99.0   \\
Grid       & 75.3/89.4 & 98.1/98.9  & 99.9/\textbf{99.7}  & 99.7/99.5  & 98.6/97.6  & \textbf{100}/98.3 & 97.3/98.7  & 90.6/87.9   & 98.6/86.0   & 99.4/98.4   \\
Leather    & 92.3/98.1 & 99.9/99.7  & \textbf{100}/99.0   & 98.7/96.3  & \textbf{100}/99.1 & \textbf{100}/99.4 & \textbf{100}/99.3 & 92.2/94.2   & 98.9/93.8   & \textbf{100}/\textbf{99.7 }   \\
Tile       & 91.5/80.2 & 97.1/96.2  & \textbf{100}/99.2 & \textbf{100}/99.4 & 99.2/95.1  & 99.7/95.7  & 99.3/95.8  & 99.9/92.7   & \textbf{100}/92.3  & \textbf{100}/\textbf{99.6}    \\
Wood       & 94.5/85.3 & 98.7/86.0  & 99.5/95.5  & 98.4/96.5  & \textbf{100}/94.7 & 99.5/95.8  & 99.6/95.1  & 97.9/86.4   & 99.1/82.9   & \textbf{100}/\textbf{97.8}\\ 
Bottle     & 99.2/95.7 & 99.9/97.2  & 98.0/99.1  & 95.6/99.2  & \textbf{100}/98.6 & \textbf{100}/98.8 & \textbf{100}/98.6 & 99.7/93.9  & \textbf{100}/91.3  & \textbf{100}/\textbf{99.4}\\
Cable      & 90.3/80.2 & 97.6/97.8  & 90.9/95.2  & 92.7/95.1  & \textbf{99.9/98.8}  & 96.1/97.2  & \textbf{99.9}/98.5  & 98.7/88.8   & 94.2/86.6   & 98.9/\textbf{98.8}   \\
Capsule    & 81.4/95.2 & 97.0/\textbf{99.1}  & 91.3/88.1  & 96.9/90.2  & 97.4/98.4  & 96.1/98.7  & 98.0/99.0  & 71.9/91.8   & 95.1/89.3   & \textbf{98.0}/98.5   \\
Hazelnut   & 98.8/95.0 & \textbf{100}/98.8 & \textbf{100}/99.7 & \textbf{100}/99.7 & \textbf{100}/98.6 & \textbf{100}/99.0 & \textbf{100}/98.7 & 99.7/91.1  & \textbf{100}/89.6  & \textbf{100}/\textbf{99.7}  \\
Metal Nut  & 77.1/83.3 & 98.5/98.6  & \textbf{100}/99.6 & \textbf{100}/99.4 & \textbf{100}/98.7 & \textbf{100}/97.3 & 99.9/98.3  & 98.8/77.8   & 99.1/79.5   & \textbf{100}/\textbf{99.7}  \\
Pill       & 84.4/89.9 & 96.2/98.9  & 97.1/97.3  & 97.4/97.2  & 97.7/98.0  & 98.7/98.1  & 97.5/97.6  & 87.1/82.6   & 88.3/84.5   & \textbf{99.3}/\textbf{99.5}   \\
Screw      & 82.4/95.8 & 93.1/98.9  & 98.7/99.3  & 97.8/99.0  & 95.1/98.3  & 97.8/\textbf{99.7}  & 98.2/99.5  & 97.2/60.3   & \textbf{99.5}/54.0   & 95.9/97.5   \\
Toothbrush & 97.1/95.5 & 98.8/99.0  & \textbf{100}/97.3 & 97.9/97.3  & \textbf{100}/98.8 & \textbf{100}/99.1 & \textbf{100}/98.6 & 79.2/84.6   & 87.5/75.5   & \textbf{100}/\textbf{99.6}  \\
Transistor & 84.9/75.9 & 92.9/98.2  & 91.7/85.2  & 88.0/84.8  & \textbf{100}/98.1 & 95.5/92.3  & 99.9/96.5  & 89.1/56.0   & 88.3/79.1   & 99.7/\textbf{98.4}   \\
Zipper     & 93.7/95.3 & 97.1/\textbf{99.1}  & \textbf{100/99.1} & \textbf{100}/98.4 & 99.5/98.6  & 97.9/98.3  & 99.5/98.9  & 99.1/93.7   & 99.7/96.9   & 99.7/98.8   \\ \midrule
Average & 88.2/90.0 & 97.5/97.7  & 97.6/96.7  & 97.1/96.3  & 99.1/98.0  & 98.7/97.8  & 99.2/98.1  & 92.2/85.3   & 96.1/85.3  & \textbf{99.4}/\textbf{99.0}   \\ \bottomrule
\end{tabular}}
\caption{Anomaly Detection and Localization on MVTec\cite{bergmann2019mvtec}. Best results on Image AUROC or Pixel AUROC are highlighted in bold.}
\label{tab:image and pixel-auroc}
\end{table*}

\textbf{Implementation Details.\quad} All images in four datasets are resized to $256 \times 256$. We use layer1, layer2 and layer3 of ResNet-18\cite{he2016resnet-18} pre-trained on ImageNet to obtain feature maps with $64 \times 64 \times 64$, $128 \times 32 \times 32$ and $256 \times 16 \times 16$ scales respectively and frozen these blocks during training. The number of prototypes depends on the dataset and accounts for 10\% of the total number of normal samples in the dataset. The maximum number of iterations of k-means for each scale is set to 300. We use Adam optimizer\cite{kingma2014adam} for the parameter optimization, with an initial learning rate $10^{-4}$ and a weight decay of $10^{-2}$. $\alpha$ and $\gamma$ in focal loss is set to 0.5 and 4 respectively. $\lambda$ in the total loss is set to 5. We train for 700 epochs with a batch size of 64 consisting of 32 normal samples, 16 extended anomalies and 16 simulated anomalies to ensure the diversity of anomalies.  We take the average of the top 100 anomalous pixels as the image-level anomaly score.
We compare PRN to seven unsupervised SOTA methods and two supervised SOTA methods. 
The results we report are based on the implementation provided by these methods.
The backbone of PatchCore\cite{roth2022patchcore}, RD4AD\cite{deng2022rd}, CFLOW\cite{gudovskiy2022cflow} and CFA\cite{lee2022cfa} is WideResNet50. SSPCAB\cite{ristea2022sspcab,akcay2022anomalib} replaces the penultimate convolutional layer of reconstructive encoder in DRAEM\cite{zavrtanik2021draem}. DevNet\cite{pang2021devnet} proposes that the anomaly score given by the network can be further back-propagated to the original image pixels to infer which pixels are the major contributors to the anomaly for anomaly localiaztion.  We use this approach to obtain the anomaly localization performance of DRA\cite{ding2022dra}.

\subsection{Anomaly Detection and Localization on MVTec}
\label{subsec:Anomaly Detection and Localization on MVTec}

Anomaly detection and localization results on MVTec are shown in \cref{tab:image and pixel-auroc}. Our method achieves the highest image AUROC (detection) and the highest pixel AUROC (localization) in 10 out of 15 classes. The average image AUROC results show that our method outperforms unsupervised SOTA by 0.2\% and supervised SOTA by 3.3\%. Meanwhile, for pixel AUROC, our method outperforms unsupervised SOTA by 0.9\% and supervised SOTA by 13.7\%.

\begin{table*}[]
\centering
\scalebox{0.7}{
\begin{tabular}{@{}lcccccccccc@{}}
\toprule
\multirow{2}{*}{Category}     & \multicolumn{7}{c}{Unsupervised}                                               & \multicolumn{3}{c}{Supervised} \\
\cmidrule(lr){2-8}\cmidrule(lr){9-11} 
            & KDAD\cite{salehi2021kdad}& CFLOW\cite{gudovskiy2022cflow} & DRAEM\cite{zavrtanik2021draem} & SSPCAB\cite{ristea2022sspcab} & CFA\cite{lee2022cfa} & RD4AD\cite{deng2022rd} & PatchCore\cite{roth2022patchcore}  & DevNet\cite{pang2021devnet} & DRA\cite{ding2022dra}  & Ours   \\ \midrule
Carpet     & 92.5/45.6  & \textbf{97.6}/68.3 & 92.9/65.1 & 86.4/48.6 & 93.6/57.2 & 95.4/56.5 & 95.5/62.2 & 85.8/45.7 & 92.2/52.3 & 97.0/\textbf{82.0}          \\
Grid       & 72.9/7.3   & 96.0/41.2 & \textbf{98.3/62.8} & 98.0/57.9   & 92.9/25.8 & 94.2/15.8 & 94.0/24.5 & 79.8/25.5 & 71.5/26.8 & 95.9/45.7          \\
Leather    & 97.5/26.8  & 99.2/64.5 & 97.4/\textbf{72.9} & 94.0/60.7   & 95.4/48.5 & 98.2/47.6 & 96.9/45.3 & 88.5/8.1  & 84.0/5.6  & \textbf{99.2}/69.7      \\
Tile       & 74.3/27.7  & 89.1/60.1 & \textbf{98.2}/95.2 & 98.1/96.1 & 83.3/55.9 & 85.6/54.1 & 91.3/56.2 & 78.9/52.3 & 81.5/57.6 & \textbf{98.2}/\textbf{96.5}          \\
Wood       & 76.5/24.3  & 82.8/29.0 & 90.3/74.6 & 92.8/78.9 & 85.9/49.0 & 91.4/48.3 & 87.1/49.3 & 75.4/25.1 & 69.7/22.7 & \textbf{95.9}/\textbf{82.6}          \\
Bottle     & 88.6/54.8  & 94.0/68.1 & 96.8/88.9 & 96.3/89.4 & 94.6/80.3 & 96.3/78.0 & 95.4/76.8 & 83.5/51.5 & 77.6/41.2 & \textbf{97.0}/\textbf{92.3}          \\
Cable      & 66.2/12.6  & 94.1/60.6 & 81.0/56.4 & 80.4/52.0   & 91.7/74.7 & 94.1/52.6 & 96.8/67.0 & 80.9/36.0 & 77.7/34.7 & \textbf{97.2}/\textbf{78.9}          \\
Capsule    & 90.1/10.1  & 94.0/48.8 & 82.7/39.6 & 92.5/46.4 & 93.0/48.3 & \textbf{95.5}/47.2 & 93.4/46.0 & 83.6/15.5 & 79.1/11.7 & 92.5/\textbf{62.2}          \\
Hazelnut   & 94.3/34.2  & 97.1/59.9 & \textbf{98.5}/92.6 & 98.2/93.4 & 95.2/60.0 & 96.9/60.7 & 90.9/53.2 & 83.6/22.1 & 86.9/22.5 & 97.4/\textbf{93.8}          \\
Metal Nut  & 76.9/34.1  & 91.5/88.0 & 97.0/97.0 & \textbf{97.7}/94.7 & 91.4/92.2 & 94.9/78.6 & 92.6/86.6 & 76.9/35.6 & 76.7/29.9 & 95.8/\textbf{98.0}          \\
Pill       & 86.4/20.9  & 95.2/82.0 & 88.4/47.6 & 89.6/48.3 & 95.4/81.9 & 96.7/76.5 & 94.5/75.7 & 69.2/14.6 & 77.0/21.6 & \textbf{97.2}/\textbf{91.3}          \\
Screw      & 85.2/6.1   & 95.8/43.9 & 95.0/\textbf{66.5} & 95.2/61.7 & 93.5/28.7 & \textbf{98.5}/52.1 & 97.5/34.7 & 31.1/1.4 & 30.1/5.0   & 92.4/44.9          \\
Toothbrush & 87.3/18.3  & 95.3/46.3 & 85.6/45.5 & 85.5/39.3 & 86.8/55.7 & 92.3/51.1 & 94.0/37.9 & 33.5/6.7  & 56.1/4.5  & \textbf{95.6}/\textbf{78.1}          \\
Transistor & 68.1/25.8  & 82.5/67.5 & 70.4/39.0 & 62.5/38.1 & \textbf{95.1}/76.2 & 83.3/54.1 & 92.3/66.9 & 39.1/6.4  & 49.0/11.0 & 94.8/\textbf{85.6}          \\
Zipper     & 86.5/31.5  & 96.6/65.2 & \textbf{96.8/77.6} & 95.2/76.4 & 94.3/65.2 & 95.3/57.5 & 96.1/62.3 & 81.3/19.6 & 91.0/42.9 & 95.5/\textbf{77.6}         \\  \midrule
Total Average   & 82.9/25.34 & 93.4/59.6 & 91.3/68.1 & 90.8/65.5 & 92.1/60.0 & 93.9/55.4 & 93.9/56.3 & 71.4/24.4 & 73.3/26.0 & \textbf{96.1/78.6} \\ \bottomrule

\end{tabular}}
\caption{Results of the PRO and AP metrics for anomaly localization performance on MVTec\cite{bergmann2019mvtec}.}
\label{tab:pixel aupro and pixel-ap}
\end{table*}

For a comprehensive presentation of the capabilities on anomaly localization, two additional metric results, PRO and AP, are shown in \cref{tab:pixel aupro and pixel-ap}. PRN outperforms the previous unsupervised SOTA by 2.2\% and the previous supervised SOTA by 22.8\% on the PRO metric. This confirms that PRN is more effective at simultaneously localizing anomalous regions of varying sizes. The more challenging AP metric further demonstrates the excellent anomaly localization capability of PRN. A better AP score is achieved in 12 out of 15 classes and is comparable to SOTA in other classes. In terms of overall AP, our approach even outperforms unsupervised SOTA by 10.5\% and supervised SOTA by 52.6\%. This significant improvement over AP goes a long way to demonstrate that PRN is more discriminative between normal and abnormal pixels. 
We further compare the pre-trained based approaches in terms of inference time per image (second) and performance, as shown in \cref{tab:Inference time on MVTec}. All experiments were conducted on NVIDIA GeForce RTX 3090, using a uniform standard. Our approach not only gains the best performance, but also significantly reduces the inference time.
 
\begin{figure*}
  \centering
   \includegraphics[width=0.93\linewidth]{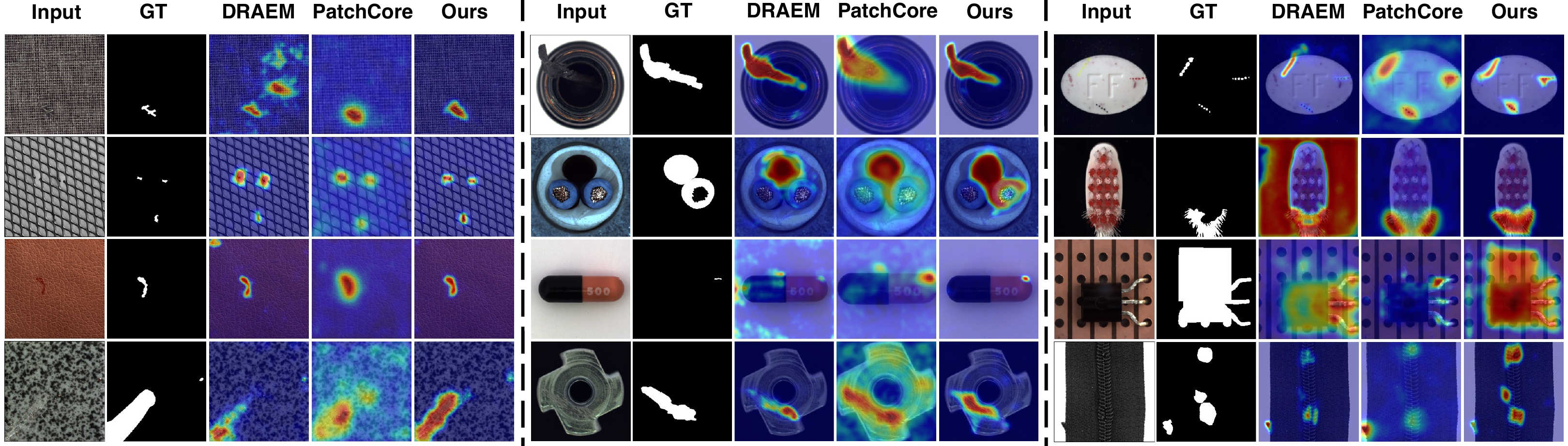}
   \caption{Qualitative examples on MVTec\cite{bergmann2019mvtec}. PRN achieves more accurate localization results for various types of anomalies. }
   \label{fig:more visualization}
\end{figure*}

We qualitatively evaluate the performance of anomaly localization compared to state-of-the-art methods DRAEM\cite{zavrtanik2021draem} and PatchCore\cite{roth2022patchcore} by visualizing the results in \cref{fig:more visualization}.  Our model accurately locates the anomalies and clearly focus on all anomalous regions, regardless of their sizes, shapes and numbers. Additional qualitative results are provided in the supplementary material.

\begin{table}[]
\centering
\scalebox{0.8}{
\begin{tabular}{@{}lcccccc@{}}
\toprule
          & Backbone                   & I $\uparrow$ & P $\uparrow$ & O $\uparrow$ & A $\uparrow$ & T $\downarrow$ \\\midrule
CFLOW     & \multirow{3}{*}{WResNet50} & 97.5          & 97.7          & 93.4          & 59.6          & 0.127          \\
RD4AD        &                            & 98.7          & 97.8          & 93.9          & 55.4          & 0.094          \\
PatchCore &                            & 99.2          & 98.1          & 93.9          & 56.3          & 0.133          \\\midrule
CFLOW     & \multirow{4}{*}{ResNet18}  & 96.2          & 98.1          & 92.8          & 59.2          & 0.106          \\
RD4AD        &                            & 97.9          & 97.1          & 92.7          & 53.7          & 0.076          \\
DRA       &                            & 96.1          & 84.1          & 71.5          & 25.7          & 0.223          \\
 \textbf{PRN(Ours)}  &                 & \textbf{99.4} & \textbf{99.0} & \textbf{96.1} & \textbf{78.6} & \textbf{0.064} \\\bottomrule
\end{tabular}}
\caption{Comparison of pre-trained based approaches in terms of performance and inference time (second) on MVTec\cite{bergmann2019mvtec}. ``I'', ``P'', ``O'', ``A'' and ``T'' respectively refer to the five metrics of image auroc, pixel auroc, pixel pro, pixel ap, and inference time per image.}
\label{tab:Inference time on MVTec}
\vspace{-1.em}
\end{table}

\subsection{Ablation Study}
\label{subsec:Ablation Study}

\textbf{The importance of MP, MSA and MF.} We investigate the importance of each modules in PRN and the results are reported in \cref{tab:ablation on module}. We have the U-Net-like architecture without any module on the skip-connection branch as the baseline. Overall, PRN outperforms the baseline by a large margin, especially on the P, O, and A metrics.
All metrics are significantly boosted by employing the MP that performs explicit residual representation. When applying the MSA which performs variable-sized anomalous feature learning, the performance is further improved. This confirms the effectiveness of information exchanging across multi-size receptive fields. Finally, removing the MF causing the degradation of performance, indicates that it is necessary to exchange information across different scales.

\begin{table}[]
\centering
\scalebox{0.8}{
\begin{tabular}{@{}cccccccc@{}}
\toprule
\multicolumn{4}{c}{Module} & \multicolumn{4}{c}{Performance} \\
\cmidrule(lr){1-4} \cmidrule(lr){5-8}
U-Net              & MP             & MSA           & MF            & I $\uparrow$      & P $\uparrow$        & O $\uparrow$        & A $\uparrow$     \\\midrule
$\checkmark$       &                &               &               & 97.4   & 91.7   & 88.6  & 58.5  \\
$\checkmark$       & $\checkmark$   &               & $\checkmark$  & 98.9   & 98.5   & 95.3  & 77.0    \\
$\checkmark$       &                &$\checkmark$   & $\checkmark$  & 97.8   & 97.0   & 92.1  & 74.0     \\
$\checkmark$       & $\checkmark$   &$\checkmark$   &               & 98.7   & 98.5   & 95.4  & 78.1     \\
$\checkmark$       & $\checkmark$   & $\checkmark$  & $\checkmark$  & \textbf{99.4}   & \textbf{99.0}     & \textbf{96.1}  & \textbf{78.6}\\\bottomrule
\end{tabular}
}
\caption{Ablations of different modules in PRN.}
\label{tab:ablation on module}
\end{table}

\textbf{Effects of different anomaly generation strategies.} We perform ablation studies to investigate the impact of the different components of the proposed anomaly generation strategies in \cref{tab:Ablation on synthetic.}. The proposed EA alleviates the problem of seen appearance variance, but does not adequately explore the underlying unseen anomalies.  \cref{tab:Ablation on synthetic.} indicates that the performance of the model increases with the variety of generated anomalies. We argue that the proposed SA consisting of both HEA and HOA can generate anomalies of various sizes, shapes and numbers, allowing our model to generalize to unseen anomalies. Besides, the proposed TA imposes soft constraints on the locations where anomaly regions are imposed, making the generated anomalies as realistic and reasonable as possible, thus significantly improving the performance of the model.

\begin{table}[]
\centering
\scalebox{0.8}{
\begin{tabular}{@{}cccccccc@{}}
\toprule
\multicolumn{4}{c}{Anomaly Generation} & \multicolumn{4}{c}{Performance} \\
\cmidrule(lr){1-4}\cmidrule(lr){5-8}
EA                  & HEA                  & HOA                & TA                      & I $\uparrow$      & P $\uparrow$        & O $\uparrow$        & A $\uparrow$     \\\midrule
$\checkmark$       &                     &                    & $\checkmark$              & 98.6    & 97.2   & 93.4   & 75.7   \\
$\checkmark$       &  $\checkmark$       &                    & $\checkmark$              & 99.1    & 98.4   & 95.4   & 77.4    \\
$\checkmark$       &                     &  $\checkmark$      & $\checkmark$              & 98.6    & 98.4   & 95.7   & 75.2   \\
                   &  $\checkmark$       &  $\checkmark$      &  $\checkmark$             & 98.7    & 98.2   & 95.1   & 73.4  \\
$\checkmark$       &  $\checkmark$       &  $\checkmark$      &                           & 98.4    & 98.4   & 94.9   & 77.6  \\
$\checkmark$       &  $\checkmark$       &  $\checkmark$      &  $\checkmark$             & \textbf{99.4}   & \textbf{99.0}     & \textbf{96.1}  & \textbf{78.6}   \\
\bottomrule
\end{tabular}}
\caption{Ablations of anomaly generation strategies.}
\label{tab:Ablation on synthetic.}
\vspace{-1.em}
\end{table}

\textbf{The effect of prototype proportion.} The effect of the ratio of prototypes to total normal samples is compared in \cref{tab:ablation on the ratio of prototypes.}. Note that $100\%$ means that no clustering is performed. Each feature map of a normal sample is regarded as a prototype and the number of prototypes is equal to the number of normal samples. The poor performance of the $\text{PRN}_{100\%}$ indicates that the residual representation obtained from the closest cluster prototype is more representative than that obtained from the single closest sample. Besides, too few prototypes lead to insufficient discrimination between prototypes, resulting in inferior performance. The results indicate that a proportion of 10\% produces the optimum performance. 
In addition, using fewer prototypes can speed up inference.

\begin{table}[]
\centering
\scalebox{0.8}{
\begin{tabular}{@{}lccccc@{}}
\toprule
             & I $\uparrow$      & P $\uparrow$        & O $\uparrow$        & A $\uparrow$    & T $\downarrow$     \\\midrule
$\text{PRN}_{5\%}$   & 99.2 & 98.6 & 95.4 & 78.1 & \textbf{0.063} \\
$\text{PRN}_{10\%}$  & \textbf{99.4} & \textbf{99.0} & \textbf{96.1} & \textbf{78.6} & 0.064 \\
$\text{PRN}_{20\%}$  & 99.2 & 98.8 & 95.7 & 77.3 & 0.066 \\
$\text{PRN}_{100\%}$ & 86.2 & 91.4 & 75.4 & 49.9 & 0.074 \\ \bottomrule
\end{tabular}}
\caption{Ablations of the ratio of prototypes to total normal samples.}
\label{tab:ablation on the ratio of prototypes.}
\end{table}

\textbf{Effects of the number of seen anomalies used.} As shown in \cref{tab:ablation on anomalies used.}, we explore the impact of the number of anomalies used. Our approach significantly outperforms Devnet\cite{pang2021devnet} and DRA\cite{ding2022dra} using different numbers of seen anomalies, which demonstrates the effectiveness of our proposed anomaly generation strategies and the robustness of PRN to datasets of different levels of imbalance.

\begin{table}[]
\centering
\scalebox{0.6}{
\begin{tabular}{@{}lcccccccccccc@{}}
\toprule
\multicolumn{1}{c}{} &\multicolumn{4}{c}{DevNet~\cite{pang2021devnet}} & \multicolumn{4}{c}{DRA~\cite{ding2022dra}}         & \multicolumn{4}{c}{\textbf{PRN(Ours)}}         \\ 
\cmidrule(lr){2-5}\cmidrule(lr){6-9} \cmidrule(lr){10-13}
\multicolumn{1}{c}{} & I $\uparrow$      & P $\uparrow$        & O $\uparrow$        & A $\uparrow$         & I $\uparrow$      & P $\uparrow$        & O $\uparrow$        & A $\uparrow$     & I $\uparrow$      & P $\uparrow$        & O $\uparrow$        & A $\uparrow$   \\\midrule
1                    &79.6  &75.3   &51.0   &16.5       & 88.9    & 78.8    & 58.2  & 19.1  & 98.8    & 98.3    & 95.4 & 74.7 \\
5                    &86.7  &83.7   &66.9   &22.7       & 93.5    & 82.8    & 68.6  & 21.9  & 99.2     & 98.6    & 95.6 & 76.4 \\
10                   &92.2  &85.3   &71.4   &24.4       & 96.1    & 85.3    & 73.3  & 26.0  & \textbf{99.4}    & \textbf{99.0}      & \textbf{96.1} & \textbf{78.6} \\ \bottomrule
\end{tabular}}
\caption{Impact of the number of seen anomalies used.}
\label{tab:ablation on anomalies used.}
\vspace{-1.25em}
\end{table}

\subsection{Evaluation on other benchmarks}
\label{subsec:Evaluation on other benchmarks}
To further evaluate the anomaly detection and localization capabilities of PRN, we benchmark PRN on three additional widely used datasets, namely DAGM\cite{wieler2007dagm}, BTAD\cite{mishra2021btad} and KolektorSDD2\cite{bozic2021SDD2}. As shown in \cref{tab:dagm_btad_sdd2}, PRN achieves new SOTA performance on all three datasets, proving its effectiveness and generalization. Results for more detailed comparisons and some qualitative examples are provided in the supplementary material.

\begin{table}[]
\centering
\scalebox{0.58}{
\begin{tabular}{@{}lcccccccccccc@{}}
\toprule
          & \multicolumn{4}{c}{DAGM\cite{wieler2007dagm}}                                      & \multicolumn{4}{c}{BTAD\cite{mishra2021btad}}      & \multicolumn{4}{c}{KolektorSDD2\cite{bozic2021SDD2}}                                \\ \cmidrule(lr){2-5}\cmidrule(lr){6-9} \cmidrule(lr){10-13}
          & I $\uparrow$      & P $\uparrow$        & O $\uparrow$        & A $\uparrow$        & I $\uparrow$        & P $\uparrow$        & O $\uparrow$        & A $\uparrow$            & I $\uparrow$        & P $\uparrow$        & O $\uparrow$        & A $\uparrow$ \\\midrule
DRAEM     & 91.1    & 83.4      & 70.5      & 35.6      & 89.0      & 87.1      & 61.6      & 19.2          &81.1       &85.6       &67.9       &39.1\\
CFLOW     & 91.2    & 95.1      & 87.6      & 45.2      & 90.5      & 96.1      & 71.6      & \textbf{54.0} &95.2       &97.4       &93.8       &46.0\\
SSPCAB    & 90.4    & 84.5      & 71.9      & 33.9      & 88.3      & 83.5      & 54.1      & 13.0          &83.4       &86.2       &66.1       &44.5\\
RD4AD        & 90.7    & 94.1      & 85.5      & 40.8      & 94.4      & 96.9      & 75.8      & 53.5          &96.0       &\textbf{97.6}       &94.7       &43.5\\
PatchCore & 92.5    & 96.1      & 88.0      & 49.0      & 92.6      & 96.9      & 76.3      & 51.5          &94.6       &97.1       &89.3       &49.8\\
DRA       & 93.5    & 95.1      & 88.8      & 47.6      & 94.2      & 75.4      & 56.2      & 12.4          &86.8       &84.4       &56.9       &3.6\\
\textbf{Ours}    & \textbf{98.2} & \textbf{96.6} & \textbf{93.8} & \textbf{49.4} & \textbf{94.7} & \textbf{97.1} & \textbf{78.0} & \textbf{54.0}  &\textbf{96.4}  &\textbf{97.6} &\textbf{94.9} &\textbf{72.5}\\ \bottomrule
\end{tabular}}
\caption{Comparison of PRN with other approaches on DAGM, BTAD, and KolektorSDD2.}
\label{tab:dagm_btad_sdd2}
\vspace{-0.5em}
\end{table}

\vspace{0.5em}
\section{Conclusion}
\label{sec:conclusion}
In this paper, we proposed a novel framework called Prototypical Residual Network for anomaly detection and localization. PRN learns residual resentations across multi-scale feature maps and within multi-size receptive fields at each scale, enabling accurate detection and localization of anomalous regions that come in a variety of sizes, shapes and numbers. In addition, we propose various anomaly generation strategies to expand and diversify the anomalies. We conduct in-depth experiments on four popular datasets to confirm the effectiveness and generalizability of our approach. PRN achieves new SOTA on anomaly detection and and significantly surpasses previous arts in anomaly localization performance.

\textbf{Limitations.\quad}Our approach requires the dataset to provide accurate ground truth masks for anomalies. Using a single image-level anomaly average score for anomalous images with different defect sizes does not favor tiny defects. We leave this intriguing extension to future work.

\vspace{0.05in}

\noindent \textbf{Acknowledgement} This project was supported by NSFC under Grant No. 62102092 and No. 62032006.

\ifarxiv \clearpage

\appendix
\label{sec:appendix}
\section{Appendix}

\subsection{Anomaly Generation Strategies}
\label{subsec:anomaly generation strategies}
This section details the generation of simulated anomalies, as shown in \cref{fig:simulated anomalies}. A noise image is generated by a Perlin noise generator\cite{perlin1985perlin,zavrtanik2021draem} (\cref{fig:simulated anomalies},$P$), and then the noise parts within a target area are retained as the ground truth mask (\cref{fig:simulated anomalies},$M$). As the shape, size, and number of generated anomalous regions vary widely, we synthesize simulated anomalies (\cref{fig:simulated anomalies},$S$) as:
\begin{equation}  
	S=\bar{M} \odot N+(1-\beta)(M \odot A)+\beta\left(M \odot N\right) 
\end{equation}
where $N$ is the normal sample, $A$ is the source image of the anomaly, $\bar{M}$ is the inverse of $M$, $\odot$ is the element-wise multiplication operation, $\beta$ is the opacity parameter for better combination of abnormal and normal regions. When $A$ is an image randomly sampled from the DTD dataset\cite{cimpoi2014DTD} and is augmented ($Aug_1$, \cref{fig:extended anomalies} in Section 3.4), we define $S$ as a HEterologous Anomaly (HEA). Correspondingly, when $A$ is an image randomly sampled from augmented normal samples, we define $S$ as a HOmology Anomaly (HOA). In particular, the normal image is first augmented ($Aug_1$, \cref{fig:extended anomalies} in Section 3.4), then is evenly divided into an $8\times8$ grid and randomly arranged before being reassembled\cite{yang2022memseg}.

\cref{fig:more anomalies} shows the anomalies generated by different strategies. In addition to increasing the number, extended anomalies (EA) increase the variety of seen anomalies. HEA and HOA supplement potential unseen anomalies with anomalies significantly different from seen anomalies.

\subsection{Dataset Split}
MVTec AD\cite{bergmann2019mvtec} is a widely used anomaly detection and localization benchmark with 15 classes, each containing one to several subclasses of anomalies. Following the general setting proposed by DRA\cite{ding2022dra}, the 10 labeled anomaly samples are sampled from all possible anomaly classes in the test set per dataset. These sampled anomalies are then removed from the test data. Both BTAD\cite{mishra2021btad} and KolektorSDD2\cite{bozic2021SDD2} are real-world industrial datasets containing three product types and one product type, respectively. The general setting used in BTAD and SDD2 is same to that used in MVTec. DAGM\cite{wieler2007dagm} contains 10 texture classes, and the original training set for each class consists of normal and abnormal samples. For each class, we first move all anomalous samples from the original training set to the original test set, and then randomly select ten anomalous samples from the test set as part of the new training set. These sampled anomalies are then removed from the test set.

\begin{figure}
  \centering
  \includegraphics[width=1.0\linewidth]{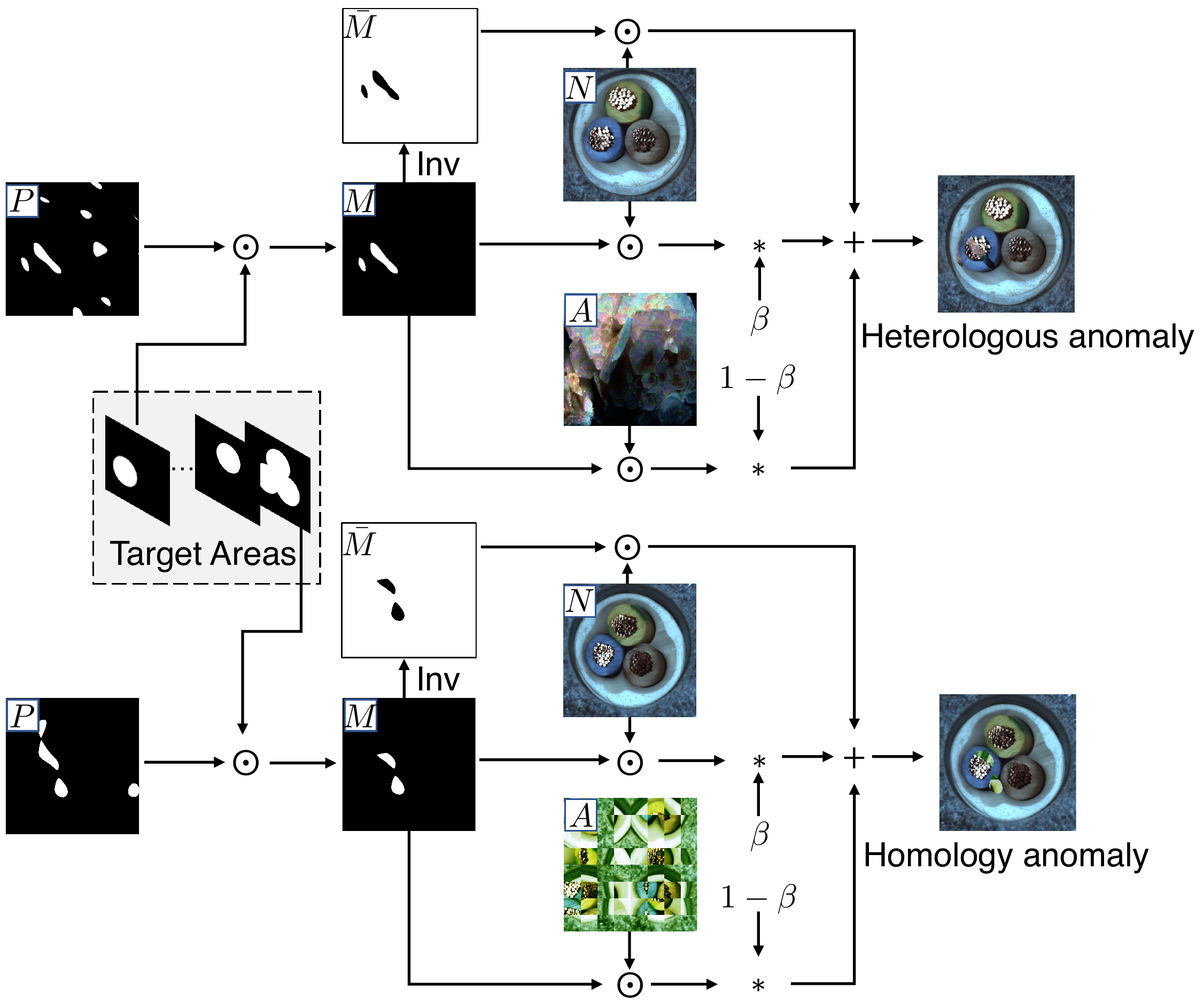}
  \caption{Generating simulated anomalies.}
  \label{fig:simulated anomalies} 
\end{figure}
\begin{figure}
  \centering
  \includegraphics[width=0.8\linewidth]{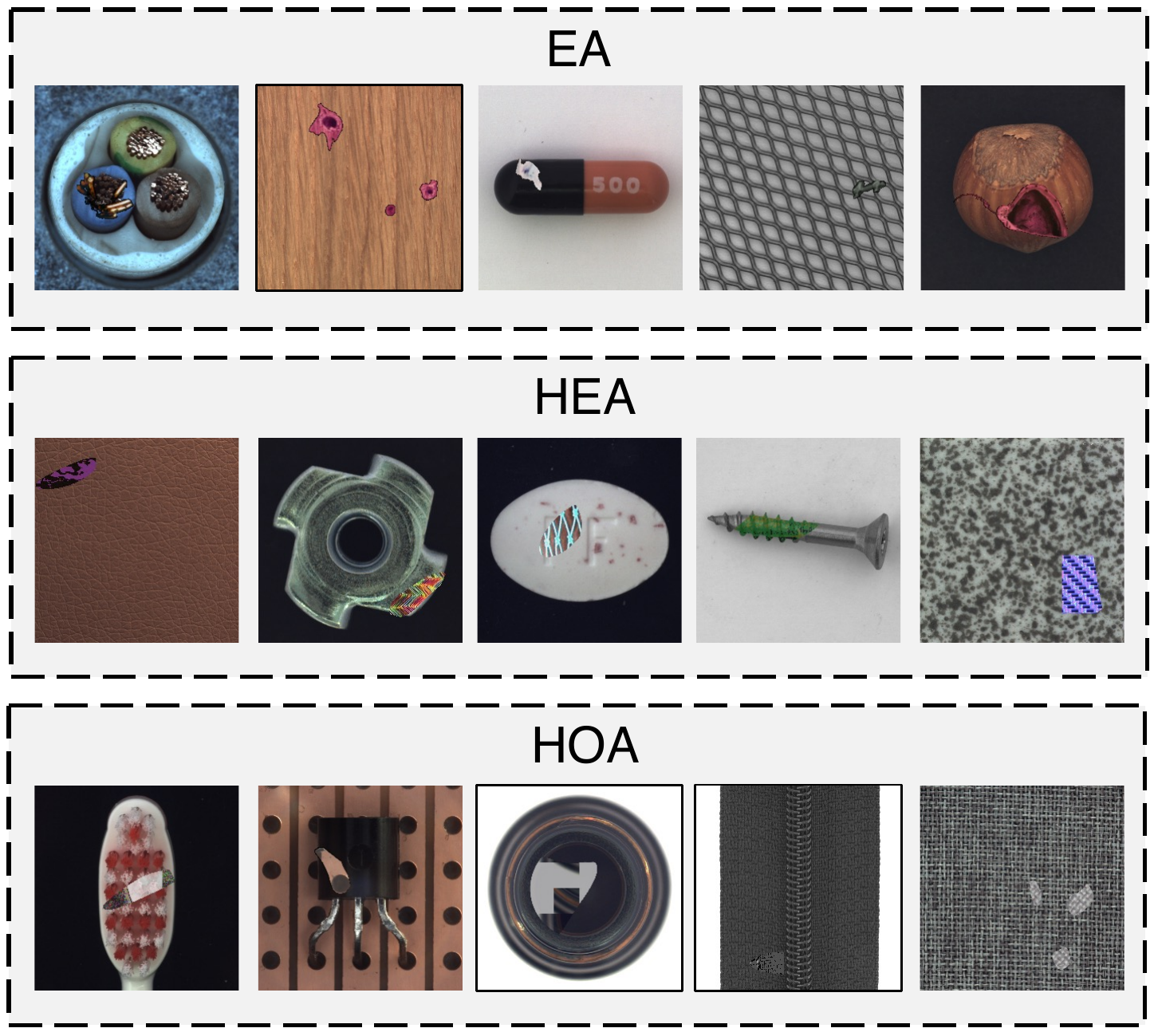}
  \caption{Examples of anomalies generated by different strategies.}
  \label{fig:more anomalies} 
\end{figure}

\begin{table*}[]
\centering
\scalebox{0.6}{
\begin{tabular}{@{}lcccccccccccccccccccccccc@{}}
\toprule
\multirow{2}{*}{Category} & \multicolumn{4}{c}{DRAEM\cite{zavrtanik2021draem}} & \multicolumn{4}{c}{CFLOW\cite{gudovskiy2022cflow}} & \multicolumn{4}{c}{SSPCAB\cite{ristea2022sspcab}} & \multicolumn{4}{c}{RD4AD\cite{deng2022rd}}    & \multicolumn{4}{c}{PatchCore\cite{roth2022patchcore}} & \multicolumn{4}{c}{\textbf{Ours}}   \\ \cmidrule(l){2-5} \cmidrule(l){6-9} \cmidrule(l){10-13} \cmidrule(l){14-17} \cmidrule(l){18-21} \cmidrule(l){22-25} 
          & I $\uparrow$      & P $\uparrow$        & O $\uparrow$        & A $\uparrow$    & I $\uparrow$      & P $\uparrow$        & O $\uparrow$        & A $\uparrow$    & I $\uparrow$      & P $\uparrow$        & O $\uparrow$        & A $\uparrow$    & I $\uparrow$      & P $\uparrow$        & O $\uparrow$        & A $\uparrow$    & I $\uparrow$      & P $\uparrow$        & O $\uparrow$        & A $\uparrow$     & I $\uparrow$      & P $\uparrow$        & O $\uparrow$        & A $\uparrow$    \\ \midrule
Class1    & 86.9 & 75.4 & 56.3 & 20.9 & 91.6 & \textbf{94.1} & 84.1 & 33.4 & 95.3  & 78.9 & 61.2 & 29.9 & 95.2 & 92.8 & 83.0   & 41.6 & 84.4  & 89.6  & 72.8  & 13.1  & \textbf{100}   & 92.7 & \textbf{90.3} & \textbf{50.1} \\
Class2    & 85.8 & 83.7 & 66.1 & 18.2 & 98.2 & 99.6 & 98.2 & 50.2 & 93.9  & 92.0   & 80.4 & 18.3 & 99.7 & \textbf{99.7} & 99.1 & \textbf{57.8} & \textbf{100}   & \textbf{99.7}  & \textbf{99.3}  & 55.5  & 96.0   & 97.1 & 95.6 & 44.8 \\
Class3    & 98.0 & 90.3 & 78.2 & 32.9 & 88.3 & 93.7 & 86.0 & 32.9 &\textbf{99.6}  & 90.3 & 79.2 & 31.7 & 81.2 & 93.9 & 85.2 & 31.8 & 94.0  & \textbf{96.2}  &\textbf{ 92.4}  & \textbf{50.9}  & 99.2 & 94.2 & 91.3 & 32.4 \\
Class4    & 99.3 & 98.6 & 95.5 & 62.4 & \textbf{100}  & \textbf{99.5} & \textbf{98.5} & 65.1 & 99.9  & 99.1 & 97.6 & 74.7 & 99.9 & 99.1 & 97.7 & 64.6 & 100   & 99.4  & 98.4  & \textbf{88.2}  & 99.7  & 98.2 & 96.7 & 67.2 \\
Class5    & \textbf{97.9} & 56.4 & 39.9 & 21.9 & 86.3 & 94.3 & 84.5 & \textbf{50.7} & 81.1  & 53.6 & 35.9 & 15.5 & 74.1 & 86.7 & 64.3 & 31.2 & 90.6  & \textbf{95.2}  & 77.3  & 29.6  & 96.9  & 94.9 & \textbf{86.1} & 30.2 \\
Class6    & \textbf{100}  & 96.0 & 89.3 & 71.5 & 96.5 & 96.1 & 87.9 & 46.9 & \textbf{100}   & 95.4 & 88.3 & 70.0 & 92.0 & 88.3 & 68.9 & 30.3 & 99.4  & 98.1  & 93.5  & 71.2  & \textbf{100}   & \textbf{98.4} & \textbf{95.7} & \textbf{71.7} \\
Class7    & \textbf{100}  & 96.7 & 90.8 & 58.1 & 98.9 & 96.0 & 91.8 & 61.4 & 100   & 94.8 & 87.0   & 51.1 & 99.8 & 95.2 & 91.4 & 65.7 & 99.9  & \textbf{96.9}  & \textbf{94.8}  & \textbf{77.7}  & \textbf{100}   & 95.1 & 91.3 & 51.3 \\
Class8    & \textbf{99.7} & 92.9 & 90.4 & 34.2 & 56.7 & 79.9 & 51.0   & 3.2  & 96.4  & 91.1 & 88.9 & 23.2 & 65.2 & 86.2 & 67.6 & 7.0    & 60.6  & 86.4  & 56.5  & 7.8   & 93.4  & \textbf{97.1} & \textbf{95.1} & \textbf{34.4} \\
Class9    & 50.2 & 49.7 & 13.3 & 0.1  & 99.9 & \textbf{99.9} & \textbf{99.8} & \textbf{65.1} & 50.9  & 60.4 & 26.1 & 0.1  & \textbf{100}  & 99.8 & 99.4 & 26.5 & 96.4  & 99.4  & 95.7  & 45.9  & 97.1  & 98.7 & 96.8 & 46.4 \\
Class10   & 92.7 & 94.2 & 85.4 & 35.7 & 95.7 & 98.0 & 94.4 & 42.9 & 86.5  & 89.1 & 74.7 & 24.4 & 99.6 & 99.0   & 97.9 & 51.1 & \textbf{99.9}   & \textbf{99.6}  & \textbf{99.0}    & 49.6  & \textbf{99.9}  & \textbf{99.6} & \textbf{99.0}   & \textbf{65.6} \\ \midrule
Average   & 91.1 & 83.4 & 70.5 & 35.6 & 91.2 & 95.1 & 87.6 & 45.2 & 90.4  & 84.5 & 71.9 & 33.9 & 90.7 & 94.1 & 85.5 & 40.8 & 92.5  & 96.1  & 88.0    & 49.0    & \textbf{98.2}  & \textbf{96.6} & \textbf{93.8} & \textbf{49.4} \\ \bottomrule
\end{tabular}}
\caption{Anomaly Detection and Localization on DAGM\cite{wieler2007dagm}. ``I'', ``P'', ``O'' and ``A'' respectively refer to the five metrics of image auroc, pixel auroc, pro and ap. The best results are highlighted in bold.}
\label{tab:dagm}
\end{table*}

\begin{table*}[]
\centering
\scalebox{0.6}{
\begin{tabular}{@{}lcccccccccccccccccccccccc@{}}
\toprule
\multirow{2}{*}{Category} & \multicolumn{4}{c}{DRAEM\cite{zavrtanik2021draem}} & \multicolumn{4}{c}{CFLOW\cite{gudovskiy2022cflow}} & \multicolumn{4}{c}{SSPCAB\cite{ristea2022sspcab}} & \multicolumn{4}{c}{RD4AD\cite{deng2022rd}}    & \multicolumn{4}{c}{PatchCore\cite{roth2022patchcore}} & \multicolumn{4}{c}{\textbf{Ours}}   \\ \cmidrule(l){2-5} \cmidrule(l){6-9} \cmidrule(l){10-13} \cmidrule(l){14-17} \cmidrule(l){18-21} \cmidrule(l){22-25} 

        & I $\uparrow$      & P $\uparrow$        & O $\uparrow$        & A $\uparrow$    & I $\uparrow$      & P $\uparrow$        & O $\uparrow$        & A $\uparrow$    & I $\uparrow$      & P $\uparrow$        & O $\uparrow$        & A $\uparrow$    & I $\uparrow$      & P $\uparrow$        & O $\uparrow$        & A $\uparrow$    & I $\uparrow$      & P $\uparrow$        & O $\uparrow$        & A $\uparrow$    & I $\uparrow$      & P $\uparrow$        & O $\uparrow$        & A $\uparrow$   \\ \midrule
01      & 98.5 & 91.5 & 61.4 & 17.0 & 93.4 & 94.8 & 60.1 & \textbf{39.6} & 96.2  & 92.4 & 62.8 & 18.1 & 98.8 & 95.7 & 72.8 & 49.3 & 96.6  & 96.5  & 78.4  & 47.1  & \textbf{100}  & \textbf{96.6} & \textbf{81.4} & 38.8 \\
02      & 68.6 & 73.4 & 39.0 & 23.3 & 79.0   & 93.9 & \textbf{56.9} & 65.5 & 69.3  & 65.6 & 28.6 & 15.8 & \textbf{84.9} & \textbf{96.0}   & 55.8 & \textbf{66.1} & 81.3  & 94.9  & 54.0    & 56.3  & 84.1   & 95.1 & 54.4 & 65.7 \\
03      & 99.8 & 96.3 & 84.3 & 17.2 & 99.1 & 99.5 & 97.9 & 56.8 & 99.4  & 92.4 & 71.0   & 5.0    & 99.5 & 99.0   & 98.8 & 45.1 & \textbf{99.9}  & 99.2  & 96.4  & 51.2  & \textbf{99.9}  & \textbf{99.6} & \textbf{98.3} & \textbf{57.4} \\ \midrule
Average & 89.0 & 87.1 & 61.6 & 19.2 & 90.5 & 96.1 & 71.6 & \textbf{54.0}   & 88.3  & 83.5 & 54.1 & 13   & 94.4 & 96.9 & 75.8 & 53.5 & 92.6  & 96.9  & 76.3  & 51.5  & \textbf{94.7} & \textbf{97.1} & \textbf{78.0}   & \textbf{54.0}   \\ \bottomrule
\end{tabular}}
\caption{Anomaly Detection and Localization on BTAD\cite{mishra2021btad}. }
\label{tab:btad}
\end{table*}

\begin{figure*}
  \centering
  \includegraphics[width=1.0\linewidth]{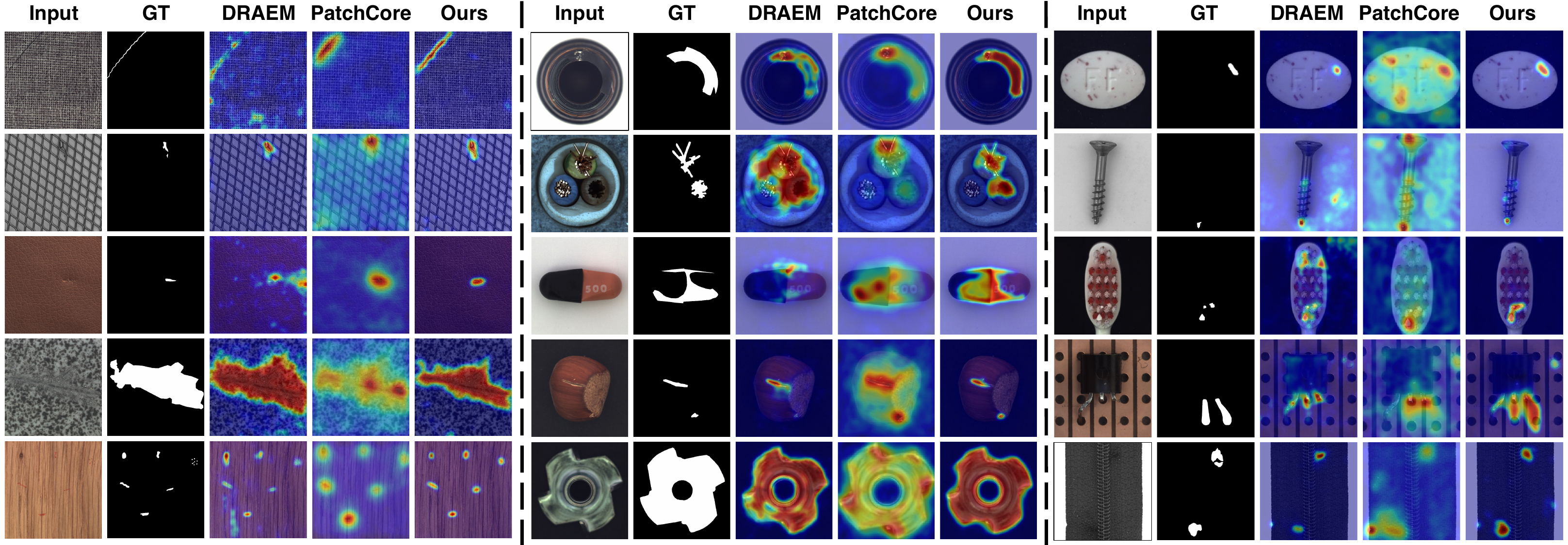}
  \caption{More qualitative examples on MVTec\cite{bergmann2019mvtec}.}
  \label{fig:mvtec supplemental visualization} 
\end{figure*}

\subsection{More Detailed Comparison}
\cref{tab:dagm} includes fine-grained anomaly detection and localization performance comparisons on all DAGM sub-datasets. We observe that PRN consistently performs well on all 10 sub-datasets and, in the average scenario, performs best across all four criteria. In particular, our approach outperforms previous methods by a large margin in two metrics, image auroc and pro. 

We also compare the anomaly detection and location performance of each method in detail on the three BTAD products and report the numerical results in \cref{tab:btad}. It can be concluded that our method achieves consistently higher performance than the others on different categories .

\begin{figure*}
  \centering
  \includegraphics[width=0.8\linewidth]{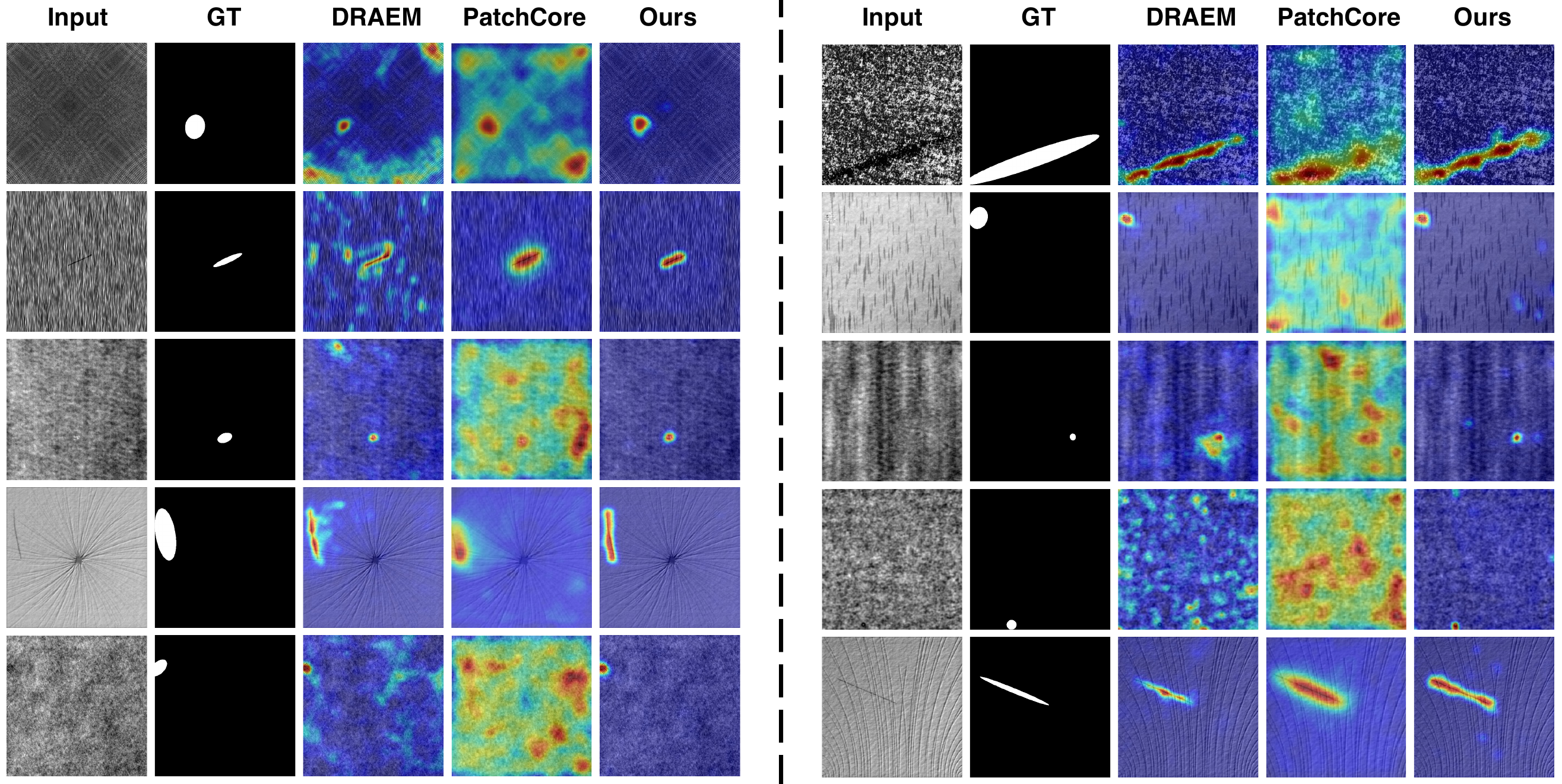}
  \caption{Qualitative examples on DAGM\cite{wieler2007dagm}.}
  \label{fig:dagm supplemental visualization} 
\end{figure*}

\begin{figure}
  \centering
  \includegraphics[width=1.0\linewidth]{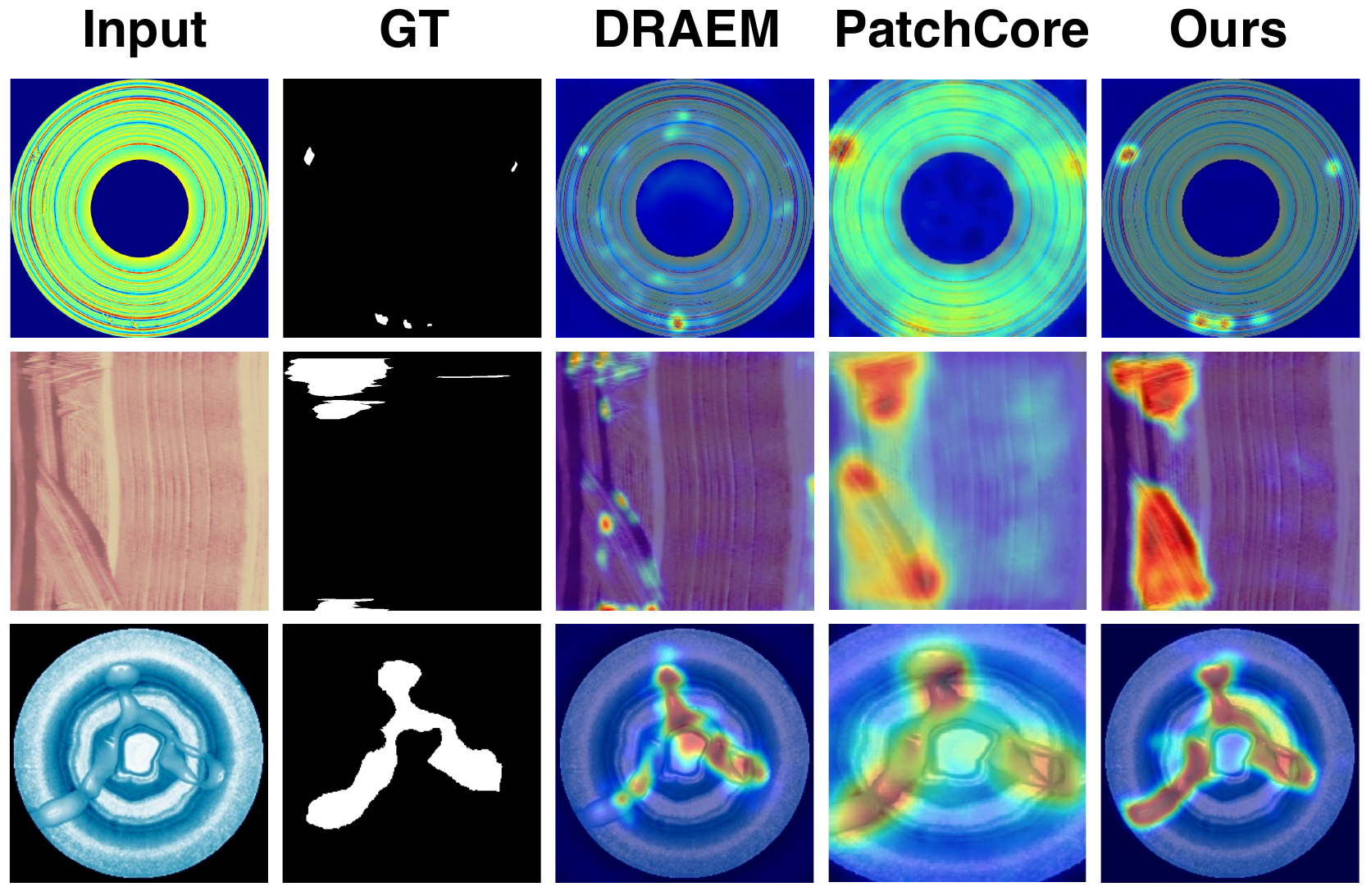}
  \caption{Qualitative examples on BTAD\cite{mishra2021btad}.}
  \label{fig:btad supplemental visualization} 
\end{figure}

\subsection{More Qualitative Examples}
We further qualitatively evaluate the performance of anomaly detection and location compared to state-of-the-art methods by introducing additional visualizations, as shown in \cref{fig:mvtec supplemental visualization}, \cref{fig:dagm supplemental visualization} and \cref{fig:btad supplemental visualization}. Our method accurately detects and localizes anomalies in a wide range of sizes, shapes and numbers, as demonstrated by qualitative comparison results. Moreover, we argue that some of the localization errors can be attributed to inaccurate ground truth labels on anomalies. An example of this is shown in the second row of \cref{fig:btad supplemental visualization}, where the ground truth does not label all anomalous regions. Another example is shown on the left in the fourth row of \cref{fig:dagm supplemental visualization}, where the ground truth labels a broad anomaly region, but our method correctly localizes the anomaly region. These imprecise annotations inevitably impact the anomaly localization scores of the evaluated methods.

\fi

{\small
\bibliographystyle{ieee_fullname}
\bibliography{PRN}
}

\end{document}

%% file: _constants.tex
\def\cvprPaperID{4728}
\def\confName{CVPR}
\def\confYear{2023}

\def\paperTitle{Prototypical Residual Networks for Anomaly Detection and Localization}

\def\authorBlock{
    Hui Zhang$^{1,2}$\qquad
    Zuxuan Wu$^{1,2}$\qquad
    Zheng Wang$^{3}$\qquad
    Zhineng Chen$^{1,2 ^*}$\qquad
    Yu-Gang Jiang$^{1,2}$
    \\
    $^{1}$Shanghai Key Lab of Intell. Info. Processing, School of CS, Fudan University \\
    $^{2}$Shanghai Collaborative Innovation Center of Intelligent Visual Computing \\
    $^{3}$School of Computer Science, Zhejiang University of Technology
    \\
}

\newif\ifreview 
\newif\ifarxiv \newcommand{\arxiv}{\arxivtrue}
\newif\ifcamera 
\newif\ifrebuttal 

%% file: cvpr_header.tex
\ifreview \usepackage[review]{cvpr} \fi
\ifarxiv \usepackage[pagenumbers]{cvpr} \fi
\ifrebuttal \usepackage[rebuttal]{cvpr} \fi
\ifcamera \usepackage{cvpr} \fi

\usepackage{graphicx}
\usepackage{amsmath}
\usepackage{amssymb}
\usepackage{booktabs}

\input{_macros}  

\usepackage{xr-hyper}

\makeatletter
\newcommand*{\addFileDependency}[1]{
  \typeout{(#1)}
  \@addtofilelist{#1}
  \IfFileExists{#1}{}{\typeout{No file #1.}}
}

\makeatother

\definecolor{R1}{HTML}{DA6A00}
\definecolor{R2}{HTML}{0BA5BE}
\definecolor{R3}{HTML}{338309}

\definecolor{citecolor}{RGB}{0, 113, 188}
\usepackage[pagebackref,breaklinks,colorlinks, citecolor=citecolor]{hyperref}

\usepackage[capitalize]{cleveref}
\crefname{section}{Sec.}{Secs.}
\crefname{table}{Table}{Tables}
\crefname{figure}{Fig.}{Figs.}

%% file: _macros.tex
\usepackage{times}
\usepackage{microtype}
\usepackage{epsfig}
\usepackage[table,xcdraw]{xcolor}
\usepackage{caption}
\usepackage{float}
\usepackage{placeins}
\usepackage{color, colortbl}
\usepackage{stfloats}
\usepackage{enumitem}
\usepackage{tabularx}
\usepackage{xstring}
\usepackage{multirow}
\usepackage{xspace}
\usepackage{url}
\usepackage{subcaption}
\usepackage{xcolor}
\usepackage[hang,flushmargin]{footmisc}

\ifcamera \usepackage[accsupp]{axessibility} \fi

% ...

\ifarxiv  \fi

\newcommand{\R}[1]{{%
    \textbf{%
        \ifstrequal{#1}{1}{\textcolor{red}{R#1}}{%
        \ifstrequal{#1}{2}{\textcolor{blue}{R#1}}{%
        \ifstrequal{#1}{3}{\textcolor{magenta}{R#1}}{%
        \ifstrequal{#1}{4}{\textcolor{teal}{R#1}}{%
                           \textcolor{cyan}{R#1}%
        }}}}%
    }%
}}